\documentclass[letterpaper]{article} % DO NOT CHANGE THIS
\usepackage[preprint]{aaai2027}  % Preserve the original manuscript layout.
\usepackage[hyphens]{url}  % DO NOT CHANGE THIS
\usepackage{graphicx} % DO NOT CHANGE THIS
\usepackage{natbib}  % DO NOT CHANGE THIS AND DO NOT ADD ANY OPTIONS TO IT
\usepackage{caption} % DO NOT CHANGE THIS AND DO NOT ADD ANY OPTIONS TO IT
\usepackage{booktabs}
\usepackage{multirow}  % 提供 \multirow
\usepackage{amssymb}
\usepackage{amsmath}
\usepackage{float}

\title{HyTBE: Hyperbolic Target-Background Expert Model for Cross-Domain Infrared Small Target Detection}
\author{
    Aohua Li\textsuperscript{\rm 1},
    Jin Kuang\textsuperscript{\rm 2},
    Yubing Lu\textsuperscript{\rm 3,\rm 4},
    Pingping Liu\textsuperscript{\rm 3,\rm 4}
}
\affiliations{
    \textsuperscript{\rm 1}College of Software, Jilin University, Changchun, Jilin 130000, China\\
    \textsuperscript{\rm 2}Hunan Engineering Research Center of Advanced Embedded Computing and Intelligent Medical Systems, Xiangnan University, Chenzhou, 423300, China\\
    \textsuperscript{\rm 3}College of Computer Science and Technology, Jilin University, Changchun, Jilin 130012, China\\
    \textsuperscript{\rm 4}Key Laboratory of Symbolic Computation and Knowledge Engineering of Ministry of Education, Jilin University, Changchun, Jilin 130012, China\\
    liah24@mails.jlu.edu.cn, gasquue@gmail.com, luyb24@mails.jlu.edu.cn, liupp@jlu.edu.cn
}

\begin{document}

\maketitle

\begin{abstract}
Infrared small target detection (IRSTD) has achieved substantial progress under domain-consistent evaluation, yet detector performance often degrades markedly when generalizing to unseen infrared domains. 
Existing methods primarily improve detection by enhancing target responses and suppressing background interference. 
However, when trained on only a limited set of source domains, their learned decision rules are inevitably established from a restricted range of source-domain target-background relation patterns. 
We formulate this cross-domain failure as target-background relation shift: unseen domains may exhibit relation patterns that are not observed during training, thereby weakening the discriminative capability learned from the source domains.
To address this problem, we propose HyTBE, a Hyperbolic Target-Background Expert model that expands source-domain relation patterns and adaptively adjusts visual representations using explicit relation cues. 
The Target-Background Relation Intervention selectively perturbs either targets or backgrounds, broadening the observable relation patterns during training while maintaining valid supervision. 
Subsequently, the Hyperbolic Relation Modeling maps multi-scale visual cues into a Poincaré ball and characterizes the target-background relation of each feature token according to its relative distances to the target and background anchors. 
The Hyperbolic-guided MoE Adapter further uses these hyperbolic relation representations to calibrate multi-scale visual features and aggregate expert-specific feature corrections for different relation patterns. 
Leave-one-domain-out experiments on NUAA-SIRST, NUDT-SIRST, and IRSTD-1K demonstrate that HyTBE achieves stronger cross-domain generalization than competitive baselines.
\end{abstract}

\begin{links}
    \link{Code}{https://github.com/PepperCS/HyTBE}
\end{links}

\begin{figure}[t]
\centering
\includegraphics[width=0.90\columnwidth]{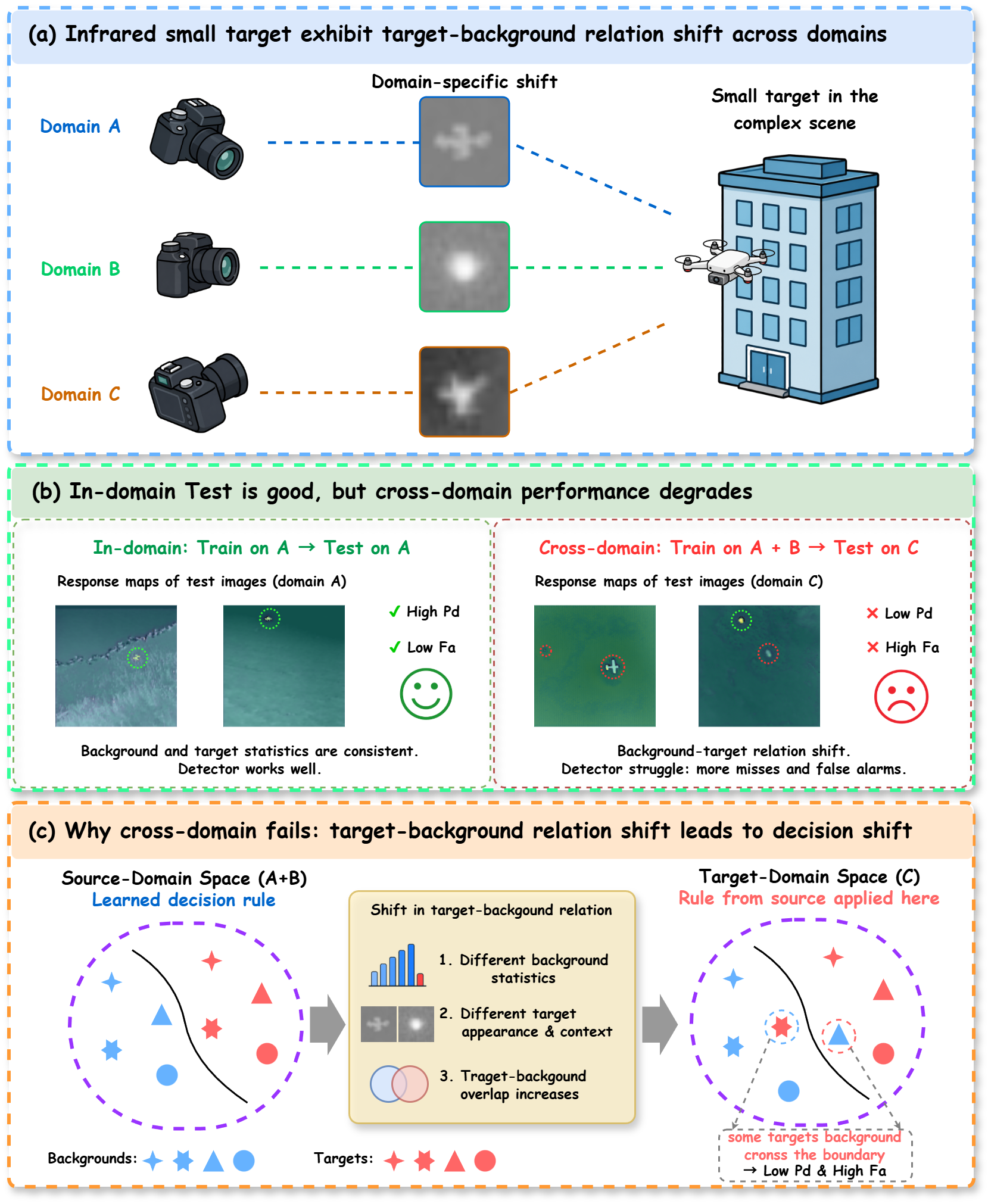}
\caption{
Motivation for cross-domain IRSTD: variations in target appearance and background statistics induce target-background relation shift, weakening source-domain decision rules in unseen domains.
}
\label{fig1}
\end{figure}

\section{Introduction}
Infrared small target detection (IRSTD) aims to locate tiny and dim targets embedded in complex infrared backgrounds \cite{kou2023infrared}, with important applications in remote sensing \cite{li2026probing}, maritime surveillance \cite{zhang2022isnet}, and early-warning systems \cite{lu2026physics}. 
Unlike generic objects \cite{huang2024dq} with recognizable textures, contours, and semantic structures, infrared small targets usually occupy only a few pixels and provide limited appearance information. 
Their weak responses can therefore be easily overwhelmed by background clutter and sensor noise. 
Recent deep-learning-based methods have substantially improved IRSTD performance through target-aware feature enhancement \cite{liu2024infrared, yang2025pinwheel}, multi-scale feature fusion \cite{wu2022uiu, li2022dense}, background suppression \cite{liu2026rpcassm}, model-driven unfolding \cite{wu2024rpcanet}, and Transformer-based representation learning \cite{liu2026pqgnet, yuan2024sctransnet}. 
These advances have enabled increasingly accurate target localization under domain-consistent evaluation. 

\textbf{\emph{Despite these advances, most existing methods have been developed and evaluated under domain-consistent settings, where the training and test data follow similar distributions.}}
In practical deployment, detectors often encounter infrared domains that are not observed during training. 
Variations in sensors, imaging platforms, environmental conditions, and scene contents can jointly alter target appearances and background statistics \cite{yuan2025ascnet, li2026ivan, duan2026cross}. 
As shown in Fig.~\ref{fig1}(a), small targets from different infrared domains may appear as compact bright spots, weak blurred responses, or irregular structures, while their surrounding backgrounds exhibit distinct textures, noise levels, and clutter distributions. 
These observation differences further lead to a clear performance discrepancy. As shown in Fig.~\ref{fig1}(b), a detector can produce concentrated target responses and effectively suppress background interference under domain-consistent evaluation, yet suffer from weakened responses, missed detections, and false alarms after being transferred to an unseen domain. 
This phenomenon indicates that a decision rule effective in the source domains does not necessarily generalize reliably to a new infrared domain.

\textbf{\emph{The central challenge of cross-domain IRSTD is that domain variations alter not only the appearances of targets and backgrounds individually, but also the discriminative relation between them.}}
Because infrared small targets contain few pixels and lack stable texture and semantic structures, their detectability depends strongly on their relative contrast, saliency, morphology, and contextual difference from the surrounding background. 
Consequently, a target that is sufficiently distinctive against one background may become ambiguous against another. 
As conceptually illustrated in Fig.~\ref{fig1}(c), a decision boundary learned from limited source-domain relation patterns may fail to reliably separate target and background features in an unseen domain. 
We refer to this variation in relation patterns and the resulting failure of the source-learned decision rule as target-background relation shift.
  
\textbf{\emph{A robust cross-domain detector should therefore both cover more relation patterns during training and adapt its features to different relations.}}
On the one hand, the source-domain training data should expose the detector to sufficiently diverse target-background relations, thereby reducing its dependence on the limited patterns contained in the original datasets. 
On the other hand, the model should explicitly characterize whether visual features are more closely associated with targets or backgrounds and adjust their representations accordingly. 
The former broadens the range of relations observed during training, whereas the latter allows the model to respond adaptively to different relation patterns. 
Together, these two requirements provide a principled path from source-domain relation diversification to unseen-domain relation generalization.

Following this principle, we propose HyTBE, a Hyperbolic Target-Background Expert model for cross-domain IRSTD. 
HyTBE first introduces Target-Background Relation Intervention, which selectively perturbs either targets or backgrounds while retaining valid supervision. 
By varying one side of the relation at a time, this strategy broadens the target-background patterns observed during source-domain training without requiring access to the target domain. 
HyTBE then performs Hyperbolic Relation Modeling, which maps multi-scale visual cues into a Poincaré ball and characterizes each feature token through its relative distances to target and background anchors. 
The hyperbolic distance metric thereby provides explicit target-oriented and background-oriented relation representations.
Finally, a Hyperbolic-guided MoE Adapter uses these relation representations to recalibrate multi-scale visual features and aggregate expert-specific feature corrections for different relation patterns. 
The three components form a progressive pipeline that diversifies target-background relations, explicitly represents them, and adapts visual features accordingly.

We evaluate HyTBE under a leave-one-domain-out protocol on three public IRSTD datasets: NUAA-SIRST, NUDT-SIRST, and IRSTD-1K. 
In each setting, two datasets are used as source domains for training, while the remaining dataset is held out as an unseen target domain without target-domain fine-tuning. 
Experimental results show that HyTBE achieves the best mIoU and F-measure across all three unseen target domains. 
Ablation studies further verify the individual contributions of Target-Background Relation Intervention, Hyperbolic Relation Modeling, and the Hyperbolic-guided MoE Adapter. 
% Visual analyses also demonstrate that HyTBE better preserves target responses and suppresses background interference under domain shifts.

The main contributions of this work are summarized as follows:
\begin{itemize}
    \item We formulate cross-domain IRSTD from the perspective of target-background relation shift, highlighting that decision rules established from limited source-domain relation patterns may become unreliable in unseen infrared domains.
    
    \item We introduce Target-Background Relation Intervention, which selectively perturbs targets or backgrounds to expand the relation patterns observed during source-domain training while retaining valid supervision.
    
    \item We develop HyTBE by combining Hyperbolic Relation Modeling with a Hyperbolic-guided MoE Adapter, allowing explicit hyperbolic target-background relation cues to guide adaptive multi-scale feature calibration for improved cross-domain generalization.
\end{itemize}

\section{Related Work}
\subsection{IRSTD under Domain Shift}
Early infrared small target detection (IRSTD) methods rely on hand-crafted priors, including local contrast \cite{bai2010analysis,chen2013local}, low-rank decomposition \cite{zhu2019infrared}, and background modeling \cite{gao2013infrared}. 
Deep networks have substantially improved in-domain performance through attention mechanisms \cite{dai2021asymmetric}, multi-scale and U-shaped architectures \cite{wu2022uiu,li2022dense,lu2026physics,zhang2022isnet}, Transformer modeling \cite{yuan2024sctransnet}, model-driven unfolding \cite{wu2024rpcanet,liu2026rpcassm}, and query-guided representations \cite{liu2026pqgnet}. 
However, these methods generally assume consistent training and test distributions, making them vulnerable to domain-specific target and background cues. Recent studies improve cross-domain robustness through semantic adaptation \cite{chi2024semantic}, perturbed training with test-time adaptation \cite{chen2024enhanced}, or frequency-domain representations \cite{fu2026rethinking}. 
Unlike these approaches, we address shifts in target-background discriminative relations through relation intervention and hyperbolic expert modeling.

\subsection{Hyperbolic Representation Learning}
Hyperbolic representation learning exploits negatively curved manifolds to represent structured relations with low distortion. 
Foundational studies introduced Poincaré embeddings \cite{nickel2017poincare} and extended neural operations to hyperbolic space \cite{ganea2018hyperbolic}. 
Subsequent works demonstrated its effectiveness in visual recognition, segmentation, contrastive learning, and domain generalization \cite{liu2020hyperbolic, atigh2022hyperbolic, ge2023hyperbolic, bi2025learning}.
Recently, hyperbolic geometry has been applied to IRSTD. 
HyperISTD improves target-background separability through Poincaré embeddings and angular constraints \cite{lu2026hyperistd}, while LoHGNet combines Lorentz encoding with high-order contextual modeling \cite{ma2026lohgnet}. 
These methods mainly employ hyperbolic geometry for feature enhancement or optimization under domain-consistent settings. 
In contrast, HyTBE characterizes each feature token through its relative hyperbolic distances to target and background anchors, and uses the resulting relation cues to guide MoE-based multi-scale feature adaptation for cross-domain IRSTD.

% \subsection{Hyperbolic Representation Learning}
% Hyperbolic representation learning models structured relations with low distortion \cite{nickel2017poincare,ganea2018hyperbolic} and has been applied to various vision tasks \cite{liu2020hyperbolic,atigh2022hyperbolic,ge2023hyperbolic,bi2025learning}. 
% Recent IRSTD methods exploit hyperbolic geometry for target-background separation and feature enhancement \cite{lu2026hyperistd,ma2026lohgnet}. 
% Unlike these methods, HyTBE models token-wise relations to target and background anchors and uses them to guide cross-domain feature adaptation. 

\subsection{Mixture-of-Experts Models}
Mixture-of-Experts (MoE) employs input-dependent routing to combine specialized
expert networks \cite{jacobs1991adaptive}. Sparse MoE enables conditional
computation \cite{shazeer2017outrageously}, with later studies improving
routing stability and load balancing \cite{fedus2022switch}. Its effectiveness
has also been demonstrated in visual recognition \cite{riquelme2021scaling},
multimodal learning \cite{mustafa2022multimodal}, and differentiable soft
expert aggregation \cite{puigcerver2024sparse}.

MoE has recently been introduced into IRSTD to capture domain-dependent
patterns \cite{duan2026cross}. Unlike domain-level routing, HyTBE uses
hyperbolic target-background relation cues to guide multi-scale MoE adapters,
enabling relation-aware feature adaptation under domain shifts.

\section{Method}
\subsection{Problem Formulation}
\subsubsection{Domain Generalization Setting.}
We study IRSTD under the leave-one-domain-out setting. 
Given multiple infrared domains $\{\mathcal{D}_1,\mathcal{D}_2,\cdots,\mathcal{D}_K\}$, the model is trained on source domains $\mathcal{D}_s=\{\mathcal{D}_k\}_{k\neq t}$ and directly evaluated on an unseen target domain $\mathcal{D}_t$. 
During training, images and annotations from $\mathcal{D}_t$ are not accessible. 
The detector predicts a response map $\hat{y}=f_\theta(x)$ for each infrared image $x$, where high-response regions indicate potential small targets.

\subsubsection{Target-Background Relation Shift.}
Different from general object detection, IRSTD relies heavily on the discriminative relation between tiny targets and their surrounding backgrounds. 
Since infrared small targets occupy only a few pixels and lack stable texture or semantic structures, their detectability is determined not only by target appearance, but also by local saliency, target morphology, background clutter, and imaging style. 
We denote such target-background relation as
\[
r=\Phi(x_t,x_b),
\]
where $x_t$ and $x_b$ represent the target candidate and its surrounding background context, respectively. 
A model trained on source domains learns a decision rule from the source relation distribution $P_s(r,y)$. 
However, in an unseen target domain, target appearance and background context may change jointly, leading to
\[
P_s(r,y) \neq P_t(r,y).
\]
As a result, the source-domain decision rule may become unreliable, causing missed detections and false alarms.

\textbf{Therefore, the key to cross-domain IRSTD lies in learning a robust target-background discrimination rule that remains reliable under relation shifts.}
This motivates us to move beyond source-domain target enhancement and explicitly consider how target-background relations vary and can be adapted across infrared domains.

\begin{figure}[t]
\centering
\includegraphics[width=0.90\columnwidth]{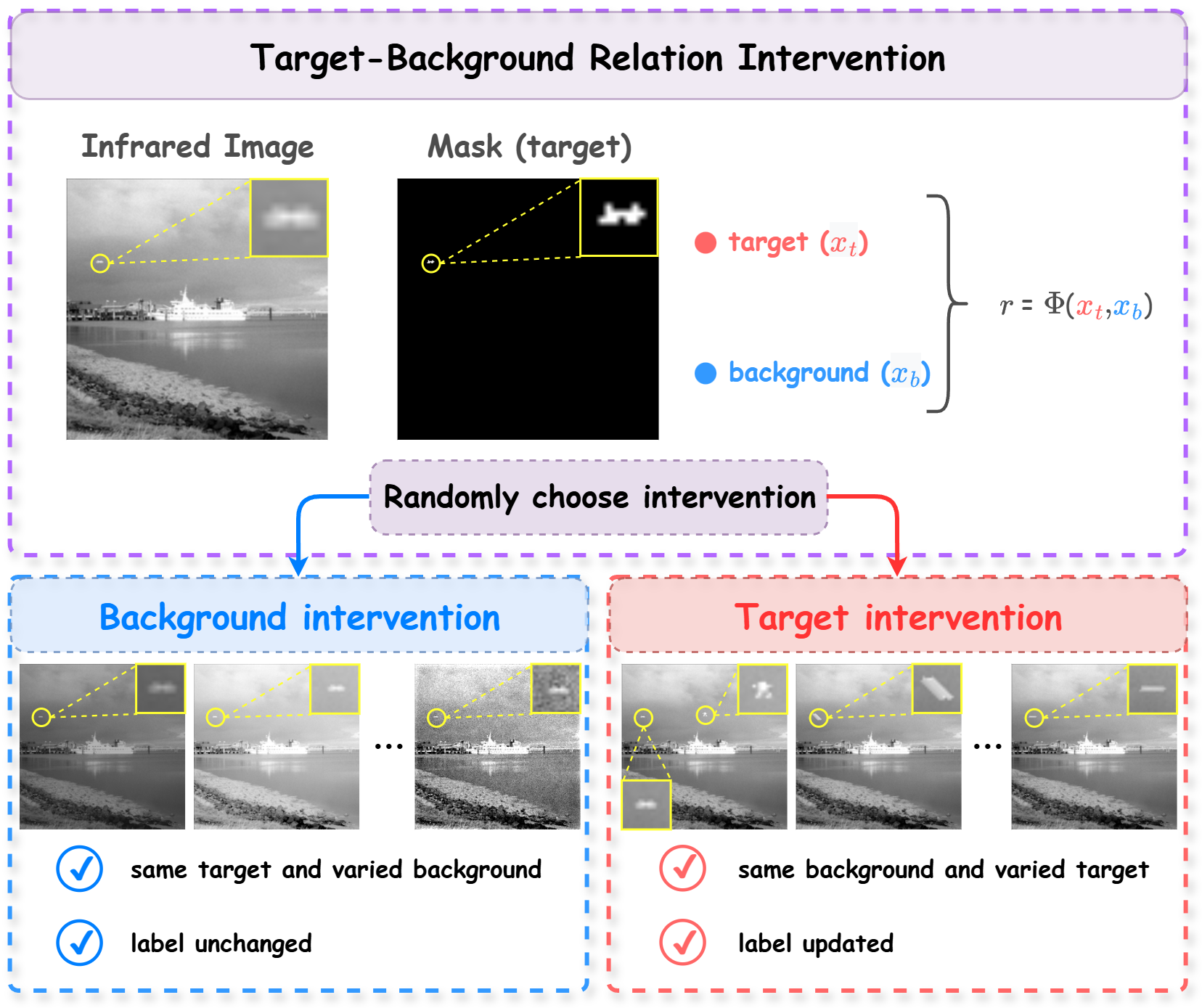}
\caption{Target-Background Relation Intervention (TBRI) selectively perturbs targets or backgrounds to diversify relation patterns while retaining valid supervision.}
\label{fig2}
\end{figure}

\subsection{Target-Background Relation Intervention}
To improve robustness to unseen domains, we introduce Target-Background Relation Intervention (TBRI) during source-domain training. 
Unlike conventional image-level augmentation, TBRI selectively modifies either the target or its background context while maintaining consistent supervision. 
As illustrated in Fig.~\ref{fig2}, TBRI consists of target intervention and background intervention, from which only one operator is selected for each intervened sample.

Given an infrared image $x$ and its binary mask $y$, the target and background
masks are defined as
\[
M_t=y, \; M_b=1-y.
\]
% For each training sample, TBRI is activated with probability $p=0.5$. Once
% activated, exactly one intervention operator is sampled:
% \[
% k\sim\operatorname{Categorical}(\boldsymbol{\pi}),
% \sum_{k=1}^{K}\pi_k=1,
% \]
% \[
% (\tilde{x},\tilde{y})=\mathcal{T}_k(x,y),
% \]
% where $\boldsymbol{\pi}=(\pi_1,\ldots,\pi_K)$ denotes the sampling probabilities and $\mathcal{T}_k$ is the selected target or background intervention operator. 
% Here, $\tilde{x}$ and $\tilde{y}$ denote the intervened image and its corresponding annotation, respectively. 
For each training sample, TBRI is activated with probability $p=0.5$. 
Once activated, exactly one intervention operator is sampled:
\[
k\sim\operatorname{Categorical}(\boldsymbol{\pi}), \;
\pi_k=\frac{1}{K}, \;
\sum_{k=1}^{K}\pi_k=1,
\]
\[
(\tilde{x},\tilde{y})=\mathcal{T}_k(x,y).
\]
where $\boldsymbol{\pi}=(\pi_1,\ldots,\pi_K)$ denotes the uniform sampling
probabilities, and $\mathcal{T}_k$ is the selected target or background
intervention operator. Here, $\tilde{x}$ and $\tilde{y}$ denote the intervened
image and its corresponding annotation, respectively.

\subsubsection{Background intervention.}
Background intervention diversifies the imaging context while preserving the
target support. 
We consider variations in global intensity response, background contrast, high-frequency components, and noise patterns. 
Let $\mathcal{B}_k(\cdot)$ denote the selected background intervention operator and $x_b^k$ its generated background candidate:
\[
x_b^k=\mathcal{B}_k(x).
\]
The intervened sample is constructed as
\[
\tilde{x}
=
M_t\odot x
+
M_b\odot x_b^k,
\,
\tilde{y}=y.
\]
Therefore, the target region and its annotation are retained, whereas the surrounding background is replaced by the perturbed candidate. 
Global imaging-style operators adjust the response of the entire image consistently without changing its binary annotation.

\subsubsection{Target intervention.}
Target intervention preserves the surrounding background while jointly transforming target appearance and target support. 
We consider variations in target saliency, brightness, morphology, and scale, together with random target sampling from the source domains.
Let $\mathcal{A}_k(\cdot)$ denote the selected target intervention operator.
It generates the transformed target response \(x_t^k\) and its corresponding binary mask \(M_t^k\):
\[
(x_t^k,M_t^k)=\mathcal A_k(x_t,M_t).
\]
The intervened sample is constructed as
\[
\tilde{x}
=
(1-M_t^k)\odot x
+
M_t^k\odot x_t^k, \;
\tilde{y}=M_t^k.
\]
Therefore, the surrounding background is preserved, whereas the target response and its corresponding annotation are jointly updated according to the selected intervention. 
For random target sampling, an additional source-domain target is transformed and inserted into a valid background location, with its support incorporated into the updated binary annotation.

\subsection{Overview of HyTBE}
\begin{figure*}[t]
\centering
\includegraphics[width=0.95\textwidth]{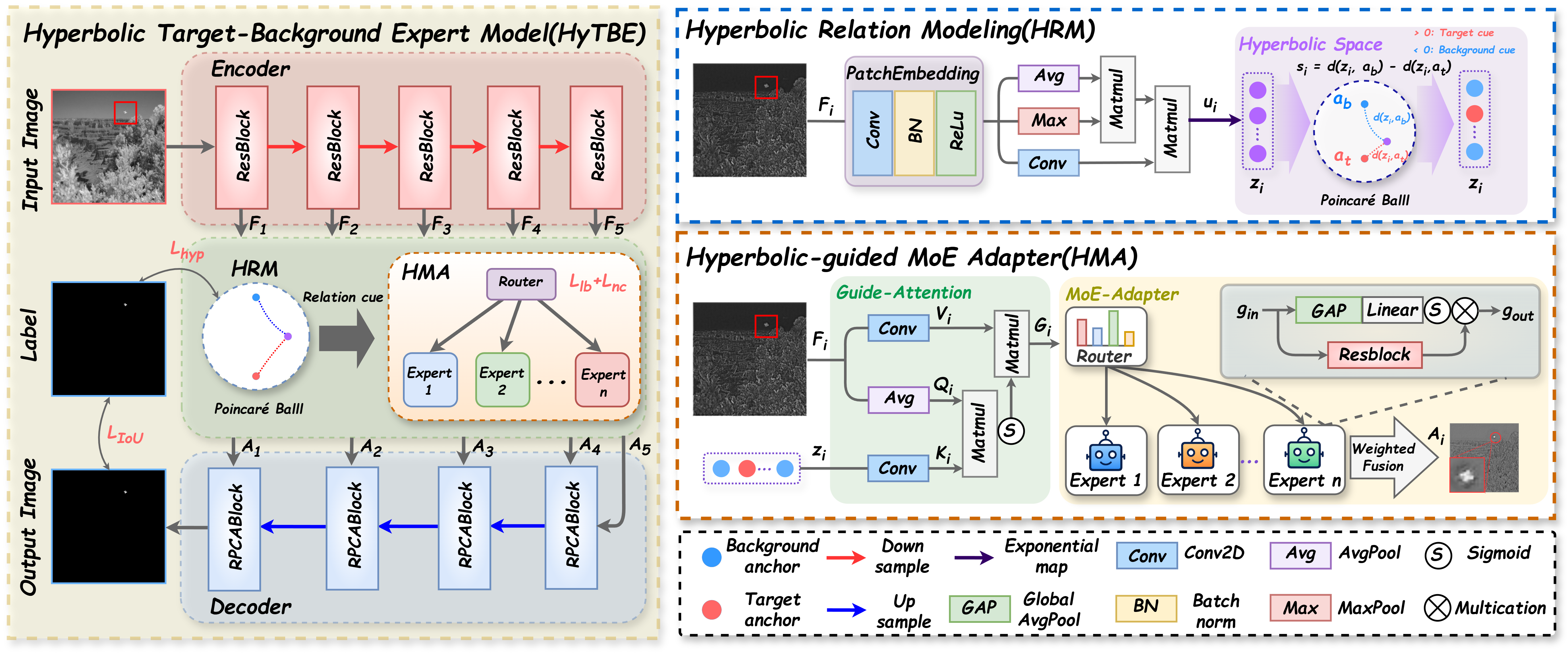} % Reduce the figure size so that it is slightly narrower than the column.
\caption{
Overall architecture of the Hyperbolic Target-Background Expert model (HyTBE). 
Its two main components are Hyperbolic Relation Modeling (HRM) for target-background relation representation and the Hyperbolic-guided MoE Adapter (HMA) for relation-guided multi-scale feature adaptation.
}
\label{fig:HYTBE}
\end{figure*}

The overall architecture of HyTBE is illustrated in Fig.~\ref{fig:HYTBE}. 
Given an infrared image $x$, an encoder composed of ResBlocks \cite{he2016deep} extracts multi-scale visual features:
\[
\{F_i\}_{i=1}^{I}=\mathcal{E}(x),
\]
where $\mathcal{E}(\cdot)$ denotes the visual encoder, $I$ is the number of
feature levels, and $F_i$ represents the visual feature at the $i$-th level.

Hyperbolic Relation Modeling (HRM) then maps the multi-scale features into
a Poincaré ball and characterizes their target-background relations according
to their relative distances to the target and background anchors:
\[
\{z_i\}_{i=1}^{I}
=
\left\{
\mathcal{F}_{\mathrm{HRM}}^{\,i}(F_i)
\right\}_{i=1}^{I},
\]
where $\mathcal{F}_{\mathrm{HRM}}^{\,i}(\cdot)$ denotes HRM at the $i$-th
level, and $z_i$ is the resulting hyperbolic relation representation.

The Hyperbolic-guided MoE Adapter (HMA) subsequently uses
$z_i$ to guide feature fusion and expert-based feature correction:
\[
\{A_i\}_{i=1}^{I}
=
\left\{
\mathcal{F}_{\mathrm{HMA}}^{\,i}(F_i,z_i)
\right\}_{i=1}^{I},
\]
where $\mathcal{F}_{\mathrm{HMA}}^{\,i}(\cdot)$ denotes HMA at the $i$-th
level, and $A_i$ represents the corresponding relation-adapted feature.

Finally, a decoder composed of RPCABlocks \cite{wu2024rpcanet,liu2026rpcassm} progressively fuses the adapted
multi-scale features and predicts the target response map:
\[
\hat{y}
=
\mathcal{D}
\left(
\{A_i\}_{i=1}^{I}
\right),
\]
where $\mathcal{D}(\cdot)$ denotes the decoder and $\hat{y}$ is
the predicted target response map. 
Overall, HyTBE follows a relation-modeling and relation-guided adaptation pipeline: 
HRM explicitly characterizes the target-background relation of visual features,
while HMA uses these relation cues to adapt feature responses to diverse
target-background patterns.

\subsubsection{Hyperbolic Relation Modeling}
Given multi-scale encoder features $\{F_i\}_{i=1}^{I}$, we employ
Patch Embedding (PE) to project features of different resolutions
onto a common spatial grid and construct a target-background relation representation:
\[
% U=\mathcal{P}\left(\{F_l\}_{l=1}^{L}\right)
% =\{u_i\}_{i=1}^{N},
u_i = \mathcal{PE} \left( F_i \right)
\]
where $\mathcal{PE}(\cdot)$ denotes PE and $u_i\in\mathbb{R}^{D}$ is the $i$-th relation token.
PE combines learned patch features with max pooled and average pooled responses to capture target saliency and background context.

Following \cite{ganea2018hyperbolic}, we map each relation token from the tangent space at the origin onto the Poincaré ball using the exponential map, where (c) is fixed to (1) throughout all experiments:
\[
z_i=\operatorname{Exp}_{0}^{c}(u_i).
\]

To establish target-background reference directions, we
construct two opposite anchors:
\[
a_t=\operatorname{Exp}_{0}^{c}(\rho),
\;
a_b=\operatorname{Exp}_{0}^{c}(-\rho),
\]
where $a_t$ and $a_b$ denote the target and background anchors, respectively,
and $\rho$ controls the anchor scale. The relation score of each token is
defined as
\[
s_i=d_c(z_i,a_b)-d_c(z_i,a_t),
\]
where $d_c(\cdot,\cdot)$ denotes the geodesic distance in the Poincaré ball \cite{ganea2018hyperbolic}.
Thus, $s_i>0$ indicates a target relation, whereas $s_i<0$ indicates
a background relation.

The ground-truth mask is resized to the token resolution using adaptive max
pooling, yielding the target and background token sets $\Omega_t$ and
$\Omega_b$. We define the target and background relation losses as
\[
\mathcal{L}_{t}
=
\frac{1}{|\Omega_t|}
\sum_{i\in\Omega_t}[m-s_i]_+
\]
\[
\mathcal{L}_{b}
=
\frac{1}{|\Omega_b|}
\sum_{i\in\Omega_b}[m+s_i]_+
\]
where $m$ is the relation margin and $[v]_+=\max(0,v)$. The complete
hyperbolic relation loss is
\[
\mathcal{L}_{\mathrm{hyp}}
=
\mathcal{L}_{t}
+
\mathcal{L}_{b}.
\]
This loss encourages target tokens to satisfy $s_i\geq m$ and background
tokens to satisfy $s_i\leq-m$, while preserving the diversity of relation
patterns within each side. 
The resulting hyperbolically constrained representation $\{z_i\}_{i=1}^{N}$ is subsequently used as the shared relation cue for HMA.

\subsubsection{Hyperbolic-guided MoE Adapter}
HMA consists of Guide-Attention and a MoE adapter. 
For the $i$-th visual feature $F_i$, its avg pooled representation provides the query, the HRM relation representation $z_i$ provides the key, and the visual feature itself provides the value:
\[
Q_i=P(F_i), K_i=\phi_k(z_i), V_i=\phi_v(F_i),
\]
where $P(\cdot)$ denotes average pooling and $\phi_k(\cdot)$ and
$\phi_v(\cdot)$ are convolutional projections. The relation-guided feature is
computed as
\[
G_i=
\operatorname{Sigmoid}
\left(
\frac{\bar Q_i\bar K_i^{\mathsf T}}{\sqrt C}
\right)V_i,
\]
where $\bar Q_i$ and $\bar K_i$ are normalized features. 
Thus, HRM relation cues recalibrate the channel responses of each encoder feature.

The guided feature is subsequently fed to a scale-specific MoE adapter. 
Each expert applies a residual block and modulates its response with a lightweight channel gate:
\[
\mathcal E_{i,e}(G_i)
=
\mathcal B_{i,e}(G_i)\otimes
\operatorname{Sigmoid}
\left(
W_{i,e}\operatorname{GAP}(G_i)
\right),
\]
where $\mathcal B_{i,e}(\cdot)$ is the residual block of the $e$-th expert,
$\operatorname{GAP}(\cdot)$ denotes global average pooling,
$W_{i,e}$ is a linear projection, and $\otimes$ multiplication.

In parallel, the router predicts input-dependent soft weights over the $E$ experts:
\[
g_i=
\operatorname{Softmax}
\left(
\operatorname{MLP}(\operatorname{GAP}(G_i))
\right).
\]
The expert corrections are then aggregated by weighted fusion and added back
to the guided feature:
\[
A_i
=
G_i+\alpha_i
\sum_{e=1}^{E}g_{i,e}\mathcal E_{i,e}(G_i),
\]
where $g_{i,e}$ is the routing weight of the $e$-th expert and $\alpha_i$ is a
learnable residual scale for the $i$-th feature level.

We employ a load-balancing loss ($\mathcal L_{\mathrm{lb}}$) \cite{fedus2022switch} to encourage balanced expert usage and a non-consistency loss ($\mathcal L_{\mathrm{nc}}$) \cite{dai2021generalizable} to promote diverse expert corrections.

% We employ a load-balancing loss $\mathcal L_{\mathrm{lb}}$ to encourage
% balanced expert usage and a non-consistency loss $\mathcal L_{\mathrm{nc}}$ to
% promote diverse expert corrections. 

% The overall objective is
% \[
% \mathcal L=
% \mathcal L_{\mathrm{IoU}}
% +\mathcal L_{\mathrm{hyp}}
% +\mathcal L_{\mathrm{lb}}
% +\mathcal L_{\mathrm{nc}}.
% \]

\subsubsection{Training Loss}
The model is jointly optimized by
$\mathcal{L}=\mathcal{L}_{\mathrm{IoU}}
+\mathcal{L}_{\mathrm{hyp}}
+\mathcal{L}_{\mathrm{lb}}
+\mathcal{L}_{\mathrm{nc}}$,
where the four terms supervise target segmentation, hyperbolic relation modeling, balanced expert routing, and expert diversity, respectively.

\begin{table*}[thb]
\centering
\caption{Cross-domain comparisons with SOTA methods on NUAA-SIRST, NUDT-SIRST, and IRSTD-1K in $mIoU$ (\%), $F$-measure (\%), $P_d$ (\%), $F_a$ ($10^{-6}$), parameters (M), and FLOPs (G).}
\label{tab1}
\resizebox{\textwidth}{!}{%
\begin{tabular}{lcccccccccccccc}
\toprule
\multirow{3}{*}{Method} &
\multicolumn{4}{c}{Source: NUAA-SIRST, NUDT-SIRST} &
\multicolumn{4}{c}{Source: NUDT-SIRST, IRSTD-1K} &
\multicolumn{4}{c}{Source: NUAA-SIRST, IRSTD-1K} &
\multirow{3}{*}{Params (M)} &
\multirow{3}{*}{FLOPs (G)} \\
\cmidrule(lr){2-5}
\cmidrule(lr){6-9}
\cmidrule(lr){10-13}
&
\multicolumn{4}{c}{Target: IRSTD-1K} &
\multicolumn{4}{c}{Target: NUAA-SIRST} &
\multicolumn{4}{c}{Target: NUDT-SIRST} &
& \\
\cmidrule(lr){2-5}
\cmidrule(lr){6-9}
\cmidrule(lr){10-13}
&
$mIoU$ & $F$ & $P_d$ & $F_a$ &
$mIoU$ & $F$ & $P_d$ & $F_a$ &
$mIoU$ & $F$ & $P_d$ & $F_a$ &
& \\
\midrule
ALCNet (TGRS 2021)
& 45.40 & 62.45 & 81.09 & 90.48
& 68.05 & 80.99 & 97.22 & 16.51
& 49.66 & 66.36 & 84.23 & 78.68
& \textbf{0.42} & \textbf{0.37} \\

DNANet (TIP 2021)
& 49.67 & 66.37 & 79.38 & \textbf{32.79}
& 68.89 & 81.58 & 94.44 & 24.59
& 51.54 & 68.02 & 81.79 & 98.88
& 4.69 & 14.26 \\

UIUNet (TIP 2022)
& \underline{51.83} & \underline{68.27} & 83.50 & 100.89
& 72.83 & 84.28 & 95.37 & 14.54
& 55.28 & 71.20 & 83.80 & \underline{43.06}
& 50.54 & 54.42 \\

MSHNet (CVPR 2024)
& 49.59 & 66.30 & 84.53 & \underline{56.17}
& 70.63 & 82.79 & 96.29 & 29.08
& 54.62 & 70.65 & 84.02 & 68.25
& 4.06 & 6.10 \\

SCTransNet (TGRS 2024)
& 48.26 & 65.10 & 84.53 & 75.07
& 70.20 & 82.49 & 96.29 & 15.61
& \underline{59.38} & \underline{74.51} & \underline{85.71} & \textbf{37.84}
& 11.19 & 10.11 \\

DRPCANet (TGRS 2025)
& 25.82 & 41.05 & 84.87 & 368.79
& 70.46 & 82.67 & 94.44 & \textbf{7.89}
& 44.64 & 61.72 & 77.67 & 158.93
& 1.16 & 73.83 \\

PConv (AAAI 2025)
& 48.14 & 64.99 & \textbf{92.09} & 126.16
& 69.28 & 81.85 & 96.29 & 14.72
& 56.75 & 72.41 & 84.65 & 53.91
& 2.93 & \underline{5.24} \\

MLPNet (TGRS 2025)
& 31.93 & 48.41 & 81.78 & 284.67
& 69.86 & 82.26 & 95.37 & 23.51
& 46.83 & 63.78 & 84.44 & 205.18
& 8.26 & 7.01 \\

PQGNet (TGRS 2026)
& 50.12 & 66.77 & 87.62 & 99.97
& \underline{73.66} & \underline{84.83} & \underline{98.69} & 58.70
& 57.66 & 73.15 & 84.12 & 55.61
& 1.20 & 9.89 \\

NS\_FPN (CVPR 2026)
& 51.18 & 67.71 & \underline{88.65} & 64.07
& 65.51 & 79.16 & 94.44 & 29.62
& 49.37 & 66.11 & 82.22 & 81.41
& 4.16 & 7.96 \\

HyTBE (Ours)
& \textbf{52.31} & \textbf{68.69} & 86.59 & 96.03
& \textbf{78.58} & \textbf{88.00} & \textbf{99.56} & \underline{13.10}
& \textbf{64.83} & \textbf{78.66} & \textbf{87.08} & 52.02
& \underline{1.13} & 8.42 \\
\bottomrule
\end{tabular}%
}
\end{table*}

\begin{table}[t]
\centering
\caption{Ablation study on the NUDT-SIRST + IRSTD-1K
$\rightarrow$ NUAA-SIRST setting.}
\label{tab:ablation}
\begin{tabular}{ccc|cccc}
\toprule
TBRI & HRM & HMA
& $mIoU\uparrow$
& $F\uparrow$
& $P_d\uparrow$
& $F_a\downarrow$ \\
\midrule
           &            &            & 68.93 & 81.60 & 95.37 & 14.89 \\
\checkmark &            &            & 71.86 & 83.63 & 97.22 & 15.61 \\
\checkmark & \checkmark &            & \underline{74.48} & \underline{85.37} & \underline{98.25} & \textbf{12.92} \\
\checkmark & \checkmark & \checkmark & \textbf{78.58} & \textbf{88.00} & \textbf{99.56} & \underline{13.10} \\
\bottomrule
\end{tabular}
\end{table}

\begin{table}[t]
\centering
\caption{Ablation study of target and background interventions in TBRI on the NUDT-SIRST + IRSTD-1K
$\rightarrow$ NUAA-SIRST setting.}
\label{tab:tbri_ablation}
\begin{tabular}{cc|cccc}
\toprule
Target & Background
& $mIoU\uparrow$
& $F\uparrow$
& $P_d\uparrow$
& $F_a\downarrow$ \\
\midrule
           &            & 71.03 & 83.06 & 97.22 & 31.05 \\
\checkmark &            & \underline{73.22} & \underline{84.54} & 96.29 & \underline{13.46} \\
           & \checkmark & 71.92 & 83.67 & \underline{98.14} & 21.36 \\
\checkmark & \checkmark & \textbf{78.58} & \textbf{88.00} & \textbf{99.56} & \textbf{13.10} \\
\bottomrule
\end{tabular}
\end{table}

\begin{table}[t]
\centering
\caption{Comparison between Euclidean and hyperbolic relation modeling on the NUDT-SIRST + IRSTD-1K $\rightarrow$ NUAA-SIRST setting. All other components and training settings are identical.}
\label{tab:geometry_ablation}
\resizebox{\columnwidth}{!}{%
\begin{tabular}{lc@{\hspace{8pt}}cccc}
\toprule
Geometry & Relation metric
& $mIoU\uparrow$
& $F\uparrow$
& $P_d\uparrow$
& $F_a\downarrow$ \\
\midrule
Euclidean & $\ell_2$ distance             & 78.08 & 87.69 & 99.03 & 17.77 \\
Hyperbolic & Poincar\'e distance           & \textbf{78.58} & \textbf{88.00} & \textbf{99.56} & \textbf{13.10} \\
\bottomrule
\end{tabular}
}
\end{table}

\begin{table}[t]
\centering
\caption{Ablation study of the HMA module on the NUDT-SIRST + IRSTD-1K
$\rightarrow$ NUAA-SIRST setting.}
\label{tab:hma_ablation}
\begin{tabular}{cc|cccc}
\toprule
Guide-Attn & MoE
& $mIoU\uparrow$
& $F\uparrow$
& $P_d\uparrow$
& $F_a\downarrow$ \\
\midrule
           &            & 74.48 & 85.37 & 98.25 & \underline{12.92} \\
\checkmark &            & 75.83 & 86.25 & \underline{99.13} & 21.90 \\
           & \checkmark & \underline{76.15} & \underline{86.46} & 98.12 & \textbf{6.82} \\
\checkmark & \checkmark & \textbf{78.58} & \textbf{88.00} & \textbf{99.56} & 13.10 \\
\bottomrule
\end{tabular}
\end{table}

\begin{table}[t]
\centering
\caption{Ablation study of the auxiliary objectives. 
$\mathcal{L}_{\mathrm{IoU}}$ is used in all variants.}
\label{tab:loss_ablation}
\resizebox{\columnwidth}{!}{
\begin{tabular}{ccccccc}
\toprule
$\mathcal L_{\mathrm{hyp}}$
& $\mathcal L_{\mathrm{lb}}$
& $\mathcal L_{\mathrm{nc}}$
& $mIoU\uparrow$
& $F\uparrow$
& $P_d\uparrow$
& $F_a\downarrow$ \\
\midrule
           &            &            & 75.47 & 86.02 & 98.07 & 26.20 \\
\checkmark &            &            & 77.18 & 87.12 & \underline{99.12} & \textbf{9.51} \\
           & \checkmark &            & 74.19 & 85.18 & 98.24 & 17.77 \\
           &            & \checkmark & 75.31 & 85.92 & 97.22 & 15.56 \\
\checkmark & \checkmark &            & 76.88 & 86.93 & 98.15 & \underline{11.43} \\
\checkmark &            & \checkmark & \underline{77.72} & \underline{87.46} & 98.84 & 11.61 \\
           & \checkmark & \checkmark & 76.18 & 86.48 & 98.16 & 17.53 \\
\checkmark & \checkmark & \checkmark & \textbf{78.58} & \textbf{88.00} & \textbf{99.56} & 13.10 \\
\bottomrule
\end{tabular}
}
\end{table}

\section{Experiments}
\subsection{Experimental Settings}
\subsubsection{Datasets.}
We conduct experiments on three widely used public infrared small target detection datasets, including NUAA-SIRST~\cite{dai2021asymmetric}, NUDT-SIRST~\cite{li2022dense}, and IRSTD-1K~\cite{zhang2022isnet}, which contain 427, 1327, and 1000 images, respectively. 
These datasets are collected from different infrared imaging scenarios and exhibit variations in target scale, target morphology, background content, and imaging style, making them suitable for evaluating cross-dataset generalization. 
For each dataset, we adopt the standard training-test split configuration used in~\cite{yuan2024sctransnet}. 
Following common IRSTD practice, pixel-level annotations are used for training and test.

\subsubsection{Cross-domain protocol.}
To evaluate cross-domain robustness, we adopt a leave-one-domain-out protocol. 
Specifically, two datasets are used as source domains for training, while the remaining dataset is held out as an unseen target domain for testing. 
This yields three cross-domain settings: NUAA-SIRST + NUDT-SIRST $\rightarrow$ IRSTD-1K, NUAA-SIRST + IRSTD-1K $\rightarrow$ NUDT-SIRST, and NUDT-SIRST + IRSTD-1K $\rightarrow$ NUAA-SIRST. 

\subsubsection{Evaluation metrics.}
Following common evaluation protocols in IRSTD, we report four widely used metrics: mean Intersection over Union ($mIoU$), F-measure ($F$), probability of detection ($P_d$), and false alarm rate ($F_a$). 
Higher mIoU, F-measure, and $P_d$ indicate better detection performance, while lower $F_a$ indicates fewer false alarms.

\subsubsection{Implementation details.}
All experiments are implemented with PyTorch on an NVIDIA GeForce RTX 4070 Ti SUPER GPU. 
The proposed HyTBE is trained from scratch without using any pretrained weights. 
For each input image, we first normalize it and then resize it to $256 \times 256$. 
Random flipping and rotation are adopted for data augmentation. 
We train the model for 200 epochs with a batch size of 4 using the Adam optimizer. 
The initial learning rate is set to 0.001.

\subsection{Comparison with State-of-the-Art Methods}
As shown in Table~\ref{tab1}, we compare HyTBE with ten representative methods, including ALCNet~\cite{dai2021attentional}, DNANet~\cite{li2022dense}, UIUNet~\cite{wu2022uiu}, MSHNet~\cite{liu2024infrared}, SCTransNet~\cite{yuan2024sctransnet}, DRPCANet~\cite{xiong2025drpca}, PConv~\cite{yang2025pinwheel}, MLPNet~\cite{wang2024mlp}, PQGNet~\cite{liu2026pqgnet}, and NS\_FPN~\cite{Yuan_2026_CVPR}.
HyTBE achieves the best mIoU and $F$-measure across all three cross-domain settings, outperforming the corresponding second-best methods by 0.48/0.42, 4.92/3.17, and 5.45/4.15 percentage points on IRSTD-1K, NUAA-SIRST, and NUDT-SIRST, respectively.
It also obtains the highest $P_d$ on NUAA-SIRST and NUDT-SIRST.
Although HyTBE does not achieve the lowest $F_a$ in every setting, its consistent improvements in mIoU and F-measure demonstrate a better overall balance between target preservation and false-alarm suppression.
Moreover, HyTBE contains only 1.13M parameters with 8.42G FLOPs, demonstrating strong cross-domain performance with a compact parameter footprint and moderate computational cost.

\subsection{Ablation Studies} 
All ablation experiments are conducted under the NUDT-SIRST + IRSTD-1K $\rightarrow$ NUAA-SIRST setting.

\textbf{Effectiveness of the main components.}
As shown in Table~\ref{tab:ablation}, the baseline achieves 68.93\% mIoU.
Introducing TBRI improves mIoU and $F$-measure by 2.93 and 2.03 percentage
points, respectively. Adding HRM further increases mIoU to 74.48\%, while the
complete model with HMA reaches 78.58\% mIoU, 88.00\% $F$-measure, and
99.56\% $P_d$. These progressive improvements verify the complementary
contributions of relation diversification, relation modeling, and
relation-guided feature adaptation.

\textbf{Effect of target-background interventions.}
Table~\ref{tab:tbri_ablation} separately evaluates the two intervention
branches while keeping the remaining architecture fixed. Target intervention
provides a larger mIoU improvement and substantially reduces $F_a$, whereas
background intervention improves $P_d$ by exposing the detector to diverse
background contexts. Combining both branches achieves the best overall
performance, improving mIoU from 71.03\% to 78.58\%. This result indicates
that target-side and background-side variations provide complementary relation
patterns.

\textbf{Geometry and HMA design.}
As reported in Table~\ref{tab:geometry_ablation}, hyperbolic modeling
consistently outperforms its Euclidean counterpart, particularly reducing
$F_a$ from 17.77 to 13.10. Table~\ref{tab:hma_ablation} further shows that
both guide-attention and MoE independently improve mIoU and $F$-measure.
Their combination achieves the best mIoU, $F$-measure, and $P_d$, confirming
that relation-guided fusion and expert adaptation work complementarily.
Although MoE alone produces the lowest $F_a$, the complete HMA provides a
better overall balance across the four evaluation metrics.

% \subsection{Analysis of Target-Background Relation Shift}

\textbf{Effect of auxiliary loss.}
As shown in Table~\ref{tab:loss_ablation}, using
$\mathcal{L}_{\mathrm{hyp}}$ alone $mIoU$ and $F$-measure by 1.71 and 1.10 percentage points, respectively, while reducing $F_a$ from 26.20 to 9.51. 
When used independently, $\mathcal{L}_{\mathrm{lb}}$ and $\mathcal{L}_{\mathrm{nc}}$ mainly reduce false alarms but do not improve the overlap-based metrics. 
Combining $\mathcal{L}_{\mathrm{nc}}$ with relation supervision further increases mIoU to 77.72\%. 
Using all three objectives achieves the best $mIoU$, $F$-measure, and $P_d$ of 78.58\%, 88.00\%, and 99.56\%, respectively, demonstrating their overall effectiveness in jointly constraining relation representation and expert adaptation.

% \subsection{Visualization}

\subsection{Visualization}
\begin{figure}[t]
\centering
\includegraphics[width=0.95\columnwidth]{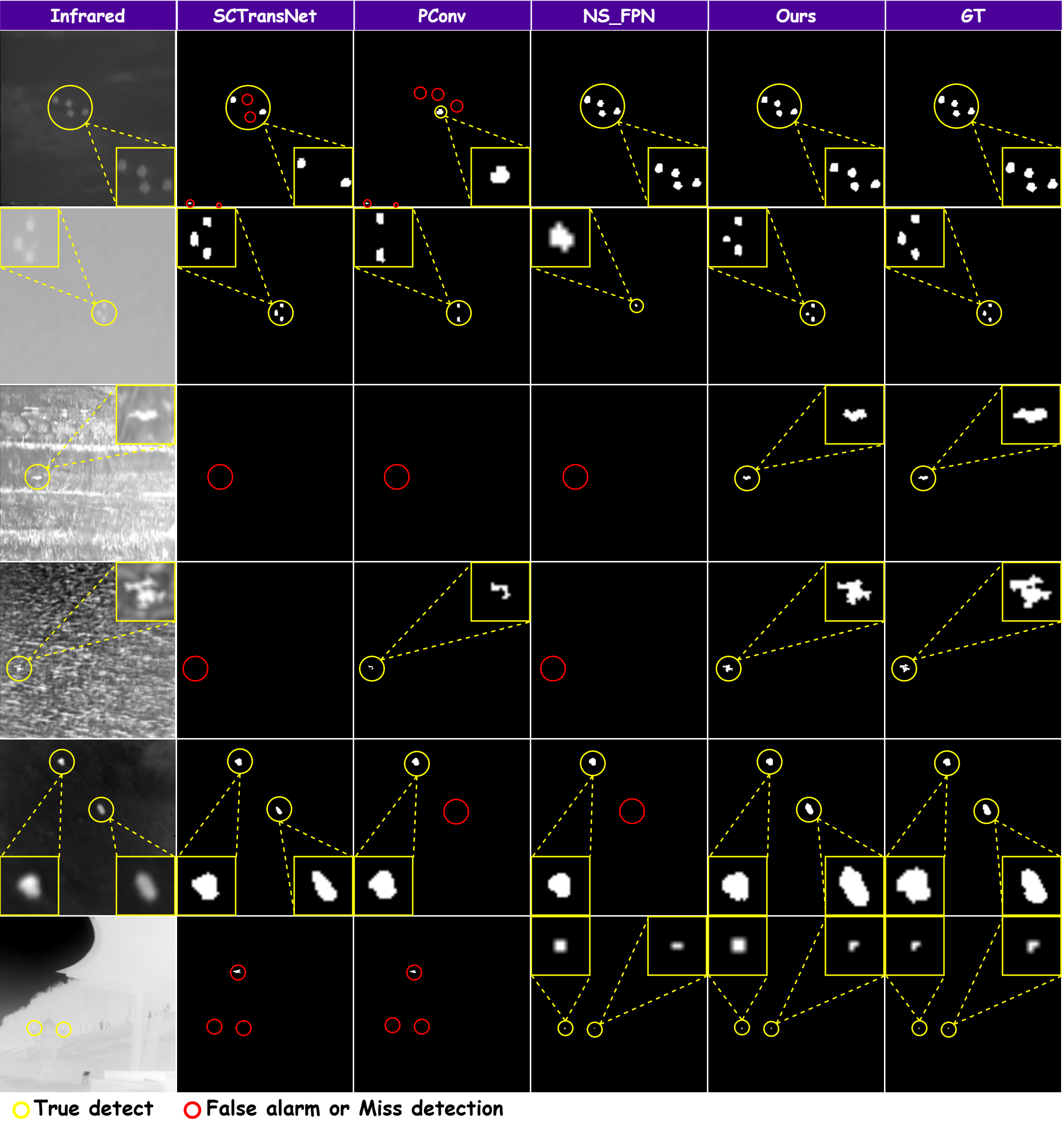}
% \caption{Qualitative cross-domain comparison with representative methods. 
% From top to bottom, each two rows correspond to NUAA-SIRST, IRSTD-1K, and NUDT-SIRST as unseen target domains, respectively. }
\caption{Qualitative cross-domain comparison with representative methods. 
From top to bottom, every two rows show results on the unseen target domains NUAA-SIRST, IRSTD-1K, and NUDT-SIRST, respectively.}
\label{fig:vis}
\end{figure}

% Fig~\ref{fig:vis} presents qualitative cross-domain comparisons on three unseen target domains. The yellow circles and enlarged boxes highlight detected targets and their local details, whereas the red circles indicate missed detections. Existing methods often fail to detect dim targets under low contrast and strong background clutter, and may produce incomplete or distorted masks. These limitations are particularly evident in multi-target scenes and cases with substantial variations in target scale and appearance. In contrast, HyTBE consistently detects targets across the three datasets while accurately preserving their number, location, and shape. Its predictions exhibit clearer target responses and closer agreement with the ground truth, demonstrating stronger robustness and generalization to unseen domains.
Fig.~\ref{fig:vis} presents qualitative comparisons on three unseen target domains. 
Existing methods often miss dim targets or produce inaccurate masks under low contrast, background clutter, and scale variations. 
In contrast, HyTBE better preserves the target number, location, and shape, producing predictions closer to the ground truth and demonstrating stronger cross-domain generalization.

\section{Conclusion}
This paper proposed a Hyperbolic Target-Background Expert model (HyTBE) to improve the cross-domain generalization of IRSTD. 
Considering that unseen infrared domains may exhibit target-background relation patterns beyond those observed during training, we introduced Target-Background Relation Intervention (TBRI) to diversify source-domain relations by selectively perturbing targets or backgrounds. 
Moreover, we developed Hyperbolic Relation Modeling (HRM) to explicitly characterize target-background relations in hyperbolic space, together with a Hyperbolic-guided MoE Adapter (HMA) that uses these relation cues to adapt multi-scale visual features. 
These three components are closely integrated to expand, represent, and adapt target-background relations for more transferable discrimination. 
Experiments across three unseen target domains demonstrate the superiority of HyTBE, highlighting its potential for robust and practical IRSTD.

\bibliography{aaai2027}
\clearpage
% Supplementary material extracted verbatim from the original standalone source.
\setcounter{secnumdepth}{2}

\twocolumn[
\begin{center}
  {\small\bfseries SUPPLEMENTARY MATERIAL\par}
  \vspace{4pt}
  {\Large\bfseries
  HyTBE: Hyperbolic Target-Background Expert Model for\\
  Cross-Domain Infrared Small Target Detection\par}
  \vspace{4pt}
  {\small Aohua Li, Jin Kuang, Yubing Lu, and Pingping Liu\par}
\end{center}
\vspace{6pt}
]

\section*{Supplementary Material Overview}

\noindent
The supplementary material is organized as follows.

\vspace{4pt}

\begin{tabular}{@{}p{0.08\columnwidth}p{0.83\columnwidth}@{}}
\toprule
\textbf{Part} & \textbf{Content} \\
\midrule
A & Evidence of Target-Background Relation Shift. \\
B & Additional details of Target-Background Relation Intervention (TBRI). \\
C & Implementation Details and Hyperparameter Analysis. \\
\bottomrule
\end{tabular}

\vspace{8pt}

\appendix

\section{Evidence of Target-Background Relation Shift}
\label{sec:supp_relation_shift}

This section provides empirical evidence for the target-background relation
shift formulated in the main paper. Given a target candidate \(x_t\) and its
surrounding background context \(x_b\), their relation is expressed as
\begin{equation}
  r=\Phi(x_t,x_b).
  \label{eq:supp_main_relation}
\end{equation}
A detector trained on the source domains establishes its decision rule from
the source relation distribution \(P_s(r,y)\). When the relation patterns in an
unseen domain are not sufficiently represented by the source domains, the
corresponding distribution changes:
\begin{equation}
  P_s(r,y)\neq P_t(r,y).
  \label{eq:supp_relation_shift_claim}
\end{equation}

Following this formulation, we organize our analysis around two questions.

\textbf{Q1: Does the target-background relation shift across domains?}

We assess its existence and magnitude through metric-wise distribution
comparisons, joint PCA visualization, and standardized Wasserstein distances.

\textbf{Q2: Is relation deviation associated with source-model degradation?}

We evaluate a fixed source-trained baseline and examine whether target
instances farther from the source-domain relation patterns exhibit higher
target-pixel miss rates. 

The following subsections address these two questions in turn.

\begin{figure}[!t]
  \centering
  \includegraphics[width=0.90\columnwidth]
  {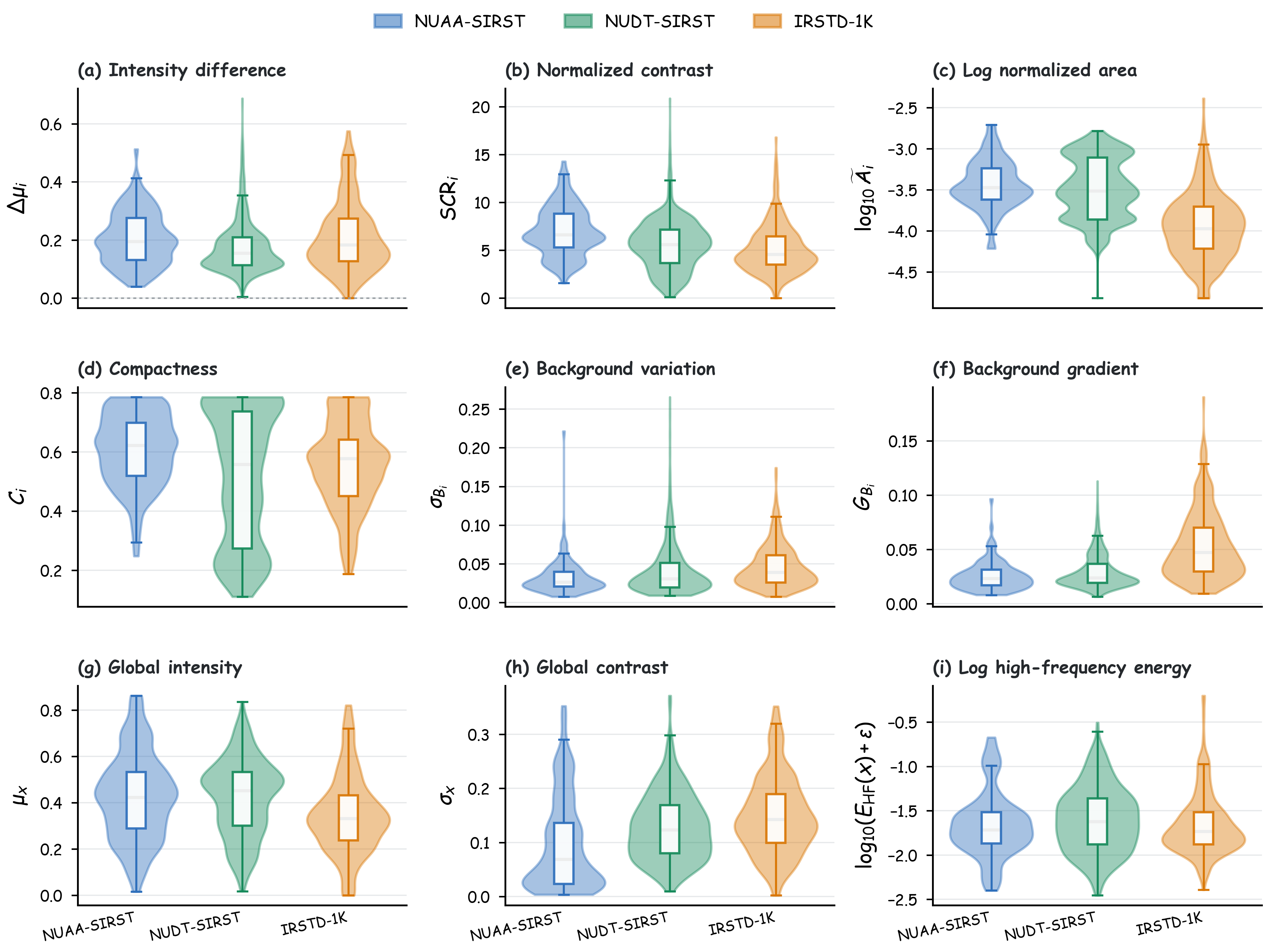}
  \caption{Cross-domain distributions of the observable target--background
  relation metrics. Panels (a)--(b), (c)--(d), (e)--(f), and (g)--(i)
  correspond to local saliency, target morphology, background clutter, and
  imaging style, respectively. Panels (c) and (i) use logarithmic mappings for
  normalized area and \(E_{\mathrm{HF}}(x)\), respectively.}
  \label{fig:supp_relation_distributions}
\end{figure}

\subsection{Analysis Protocol}
% We follow the three leave-one-domain-out settings used in the main paper. 
% In each setting, two datasets form the source domains \(\mathcal{D}_s\), and the remaining dataset is treated as the unseen target domain \(\mathcal{D}_t\).
All datasets adopt the publicly available splits from their original works, with no re-partitioning applied.

\subsubsection{Target and Local Background Extraction.}
For a consistent measurement scale, we convert all images to grayscale, scale
their intensities to \([0,1]\), and resize them to \(256\times256\); masks are
resized using nearest-neighbor interpolation.
Each connected component in the ground-truth mask is regarded as an individual target \(T_i\). 
To determine how much surrounding context should be associated with this target, we first compute its equivalent diameter
\begin{equation}
  d_i=2\sqrt{|T_i|/\pi}.
\end{equation}
We then expand the target mask outward by \(\rho_i=\lceil d_i\rceil\) pixels and define its local background as
\begin{equation}
  B_i=
  \left[
    \operatorname{Dilate}(T_i;\rho_i)\cap\Omega
  \right]
  \setminus
  \bigcup_j T_j,
  \label{eq:supp_context_region}
\end{equation}
where \(\Omega\) denotes the valid image region and \(\bigcup_jT_j\) contains all annotated targets in the image. 
Thus, \(B_i\) contains only valid background pixels around \(T_i\), without including the target itself or other nearby targets. 
The corresponding observations are \(x_{t,i}=x|_{T_i}\) and \(x_{b,i}=x|_{B_i}\). 
We use the same construction for all datasets. 
This step only constructs a target and its corresponding local background pair for subsequent measurements.

\subsubsection{Target-Background Relation Metrics.}
Following the main formulation, the relation between target \(T_i\) and its local background \(B_i\) depends on local saliency, target morphology, background clutter, and imaging style. 
Accordingly, for empirical analysis, we construct a descriptor of \(r_i=\Phi(x_{t,i},x_{b,i})\) by combining measurements of these four factors:
\begin{equation}
  r_{\mathrm{obs},i}
   =[r_{\mathrm{sal},i},r_{\mathrm{mor},i},
     r_{\mathrm{clu},i},r_{\mathrm{sty},i}],
  \label{eq:supp_observable_relation}
\end{equation}
where \(r_{\mathrm{sal},i}\) measures the intensity contrast between the target and its local background; 
\(r_{\mathrm{mor},i}\) describes the target scale and shape; 
\(r_{\mathrm{clu},i}\) measures the intensity variation and structural complexity of the local background; 
and \(r_{\mathrm{sty},i}\) summarizes the image-level intensity and frequency characteristics under which the pair is observed.
Each term is a group of metrics whose exact components are defined below. 
We use \(\epsilon=10^{-6}\) throughout.

\subsubsection{Local Saliency.}
Let \(\mu_{T_i}\) and \(\mu_{B_i}\) denote the mean intensities of target
\(T_i\) and its local background \(B_i\), respectively. We characterize local
saliency using the signed intensity difference \(\Delta\mu_i\) and its
background-normalized contrast \(\mathrm{SCR}_i\):
\begin{equation}
  r_{\mathrm{sal},i}
  =
  [\Delta\mu_i,\mathrm{SCR}_i].
  \label{eq:supp_saliency_metrics}
\end{equation}
The two components are defined as
\begin{equation}
  \Delta\mu_i
  =
  \mu_{T_i}-\mu_{B_i},
  \;
  \mathrm{SCR}_i
  =
  \frac{\Delta\mu_i}{\sigma_{B_i}+\epsilon}.
  \label{eq:supp_saliency_components}
\end{equation}
Here, the background standard deviation is
\begin{equation}
  \sigma_{B_i}
  =
  \sqrt{
    \frac{1}{|B_i|}
    \sum_{p\in B_i}
    \left(x(p)-\mu_{B_i}\right)^2
  }.
  \label{eq:supp_background_std}
\end{equation}
A positive \(\Delta\mu_i\) indicates that the target is brighter than its local
background, while a larger \(\mathrm{SCR}_i\) indicates stronger target
saliency relative to background fluctuations.

\subsubsection{Target Morphology.}
Let \(A_{T_i}=|T_i|\) and \(P_{T_i}\) denote the area and perimeter of target
\(T_i\), respectively. We characterize target morphology using normalized
area \(\widetilde A_i\) and compactness \(C_i\):
\begin{equation}
  r_{\mathrm{mor},i}
  =
  [\widetilde A_i,C_i],
  \label{eq:supp_morphology_metrics}
\end{equation}
where
\begin{equation}
  \widetilde A_i
  =
  \frac{A_{T_i}}{HW},
  \;
  C_i
  =
  \frac{4\pi A_{T_i}}{P_{T_i}^2+\epsilon},
  \label{eq:supp_morphology_components}
\end{equation}
and \(H=W=256\). Here, \(\widetilde A_i\) measures the relative target scale,
whereas \(C_i\) measures shape compactness. A larger \(C_i\) indicates a more
regular and spatially concentrated target shape.

\subsubsection{Background Clutter.}
Background clutter reflects both intensity fluctuations and spatial structures
around the target. We characterize it using the background standard deviation
\(\sigma_{B_i}\) in Eq.~\ref{eq:supp_background_std} and the average gradient
strength \(G_{B_i}\):
\begin{equation}
  r_{\mathrm{clu},i}
  =
  [\sigma_{B_i},G_{B_i}],
  \label{eq:supp_clutter_metrics}
\end{equation}
where
\begin{equation}
  G_{B_i}
    =
    \frac{1}{|B_i|}
    \sum_{p\in B_i}
    \left\|\nabla x(p)\right\|_2.
  \label{eq:supp_background_gradient}
\end{equation}
The gradient \(\nabla x\) is computed using the Sobel operator. Thus,
\(\sigma_{B_i}\) measures intensity fluctuations, whereas \(G_{B_i}\) measures
the strength of local edges and textures. Larger values indicate a less uniform
and more cluttered local background.

\subsubsection{Imaging Style.}
We characterize the image-level appearance under which each target--background
pair is observed using global intensity \(\mu_x\), global contrast \(\sigma_x\),
and high-frequency energy ratio \(E_{\mathrm{HF}}(x)\):
\begin{equation}
  r_{\mathrm{sty},i}
  =
  [\mu_x,\sigma_x,E_{\mathrm{HF}}(x)],
  \label{eq:supp_style_metrics}
\end{equation}
where \(\mu_x\) and \(\sigma_x\) are the mean and standard deviation of all
intensities in image \(x\), respectively. Let \(F_x\) be the centered Fourier
spectrum of \(x-\mu_x\). We compute the high-frequency energy ratio as
\begin{equation}
  E_{\mathrm{HF}}(x)
  =
  \frac{\sum_{(u,v)\notin\mathcal{L}}|F_x(u,v)|^2}
       {\sum_{u,v}|F_x(u,v)|^2+\epsilon},
  \label{eq:supp_high_frequency_energy}
\end{equation}
where \(\mathcal{L}\) is the centered low-frequency region whose width and
height are one quarter of the spectrum size. All targets in the same image
therefore share the same imaging-style descriptor.

\begin{figure}[t]
  \centering
  \includegraphics[width=0.90\columnwidth]{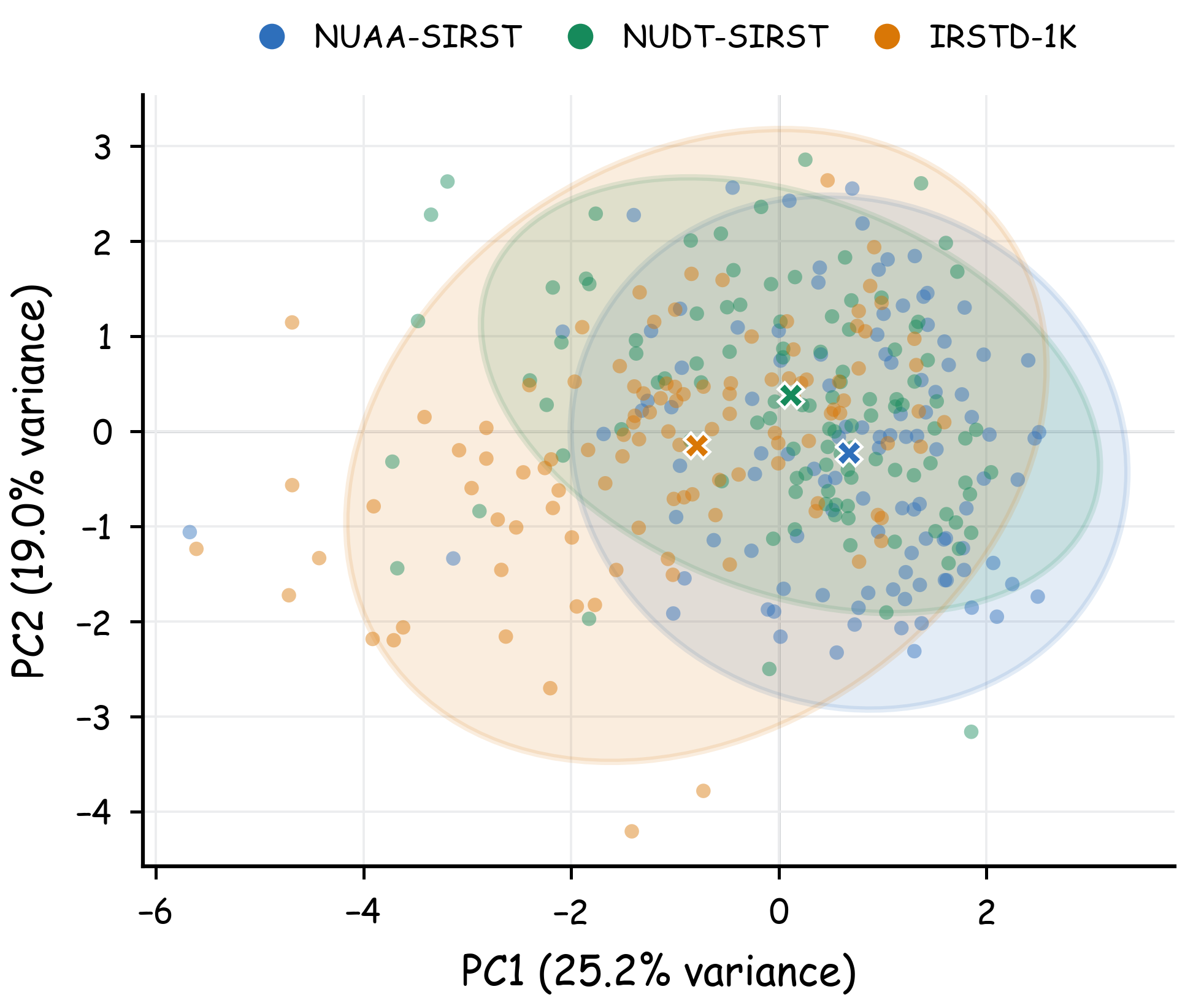}
  \caption{PCA visualization of the joint target-background relation distribution. 
  Points, crosses, and ellipses denote target-background pairs, dataset centroids, and 90\% probability regions, respectively. 
  % The plot shows 108 pairs per dataset, while all test pairs are used for PCA and the summary statistics.
  }
  \label{fig:supp_relation_joint_pca}
\end{figure}

\subsection{Relation Distributions in Unseen Domains}

\subsubsection{Marginal Distributions.}
Fig.~\ref{fig:supp_relation_distributions} shows clear cross-domain
differences across all four relation factors. NUAA-SIRST exhibits higher
target--background contrast, whereas IRSTD-1K generally contains smaller
targets and stronger local background gradients. Differences in imaging style
further show that the shift spans multiple relation properties.

\subsubsection{Joint Distribution.}
We concatenate the metrics in Eq.~\ref{eq:supp_observable_relation} into
\(\mathbf r_{\mathrm{obs},i}\) and standardize each scalar dimension as
\begin{equation}
  \widetilde r_{i,j}
  =
  \frac{r_{i,j}-\mu_j}{\sigma_j+\epsilon},
  \label{eq:supp_relation_standardization}
\end{equation}
where \(\mu_j\) and \(\sigma_j\) are estimated from an equally weighted mixture
of the three test domains. PCA is fitted with the same dataset-level weighting
to avoid dataset-size bias. The first two components explain \(44.2\%\) of the
variance. As shown in Fig.~\ref{fig:supp_relation_joint_pca}, IRSTD-1K and
NUAA-SIRST shift toward negative and positive PC1, respectively, while
NUDT-SIRST has a higher center on PC2. Despite partial overlap, these systematic
displacements provide joint-distribution evidence of relation shift.

\subsubsection{Quantitative Distribution Distance.}
For each relation-factor group \(g\), we average the standardized
1-Wasserstein distances of its constituent metrics:
\begin{equation}
  D_g(\mathcal{D}_a,\mathcal{D}_b)
  =
  \frac{1}{|g|}
  \sum_{j\in g}
  W_1
  \left(
    P_a(\widetilde{r}_j),
    P_b(\widetilde{r}_j)
  \right),
  \label{eq:supp_factor_wasserstein}
\end{equation}
where the four groups follow Eq.~\ref{eq:supp_observable_relation}; the Average
row weights them equally. Larger values indicate stronger shifts. Distances are
computed from the original standardized metrics, rather than the logarithmic
values used only for visualization. We obtain 95\% confidence intervals from
2,000 image-cluster bootstrap resamples, retaining all targets from each
resampled image.
Table~\ref{tab:relation_shift} shows the largest average discrepancy for
NUAA--IRSTD (\(0.56\)), followed by NUDT--IRSTD (\(0.47\)) and NUAA--NUDT
(\(0.34\)). Background clutter dominates both comparisons involving IRSTD-1K,
while target morphology contributes strongly to NUDT--IRSTD. Together with the
marginal and PCA results, these distances establish a systematic, multi-factor
relation shift across domains.

\begin{table}[t]
\centering
\caption{Pairwise standardized 1-Wasserstein distances between relation
descriptors. Brackets report image-cluster bootstrap 95\% confidence
intervals; larger values indicate stronger shifts.}
\label{tab:relation_shift}
\scriptsize
\setlength{\tabcolsep}{3pt}
\begin{tabular}{@{}lccc@{}}
\toprule
Relation factor & NUAA--NUDT & NUAA--IRSTD & NUDT--IRSTD \\
\midrule
Local saliency & \shortstack{0.45\\[-1pt]{[0.32, 0.61]}} & \shortstack{0.45\\[-1pt]{[0.37, 0.62]}} & \shortstack{0.33\\[-1pt]{[0.26, 0.42]}} \\
Target morphology & \shortstack{0.42\\[-1pt]{[0.38, 0.53]}} & \shortstack{0.54\\[-1pt]{[0.44, 0.68]}} & \shortstack{0.64\\[-1pt]{[0.57, 0.71]}} \\
Background clutter & \shortstack{0.23\\[-1pt]{[0.16, 0.36]}} & \shortstack{0.81\\[-1pt]{[0.63, 0.99]}} & \shortstack{0.61\\[-1pt]{[0.49, 0.75]}} \\
Imaging style & \shortstack{0.26\\[-1pt]{[0.24, 0.36]}} & \shortstack{0.43\\[-1pt]{[0.34, 0.58]}} & \shortstack{0.31\\[-1pt]{[0.24, 0.40]}} \\
\midrule
Average & \shortstack{0.34\\[-1pt]{[0.31, 0.42]}} & \shortstack{0.56\\[-1pt]{[0.50, 0.66]}} & \shortstack{0.47\\[-1pt]{[0.42, 0.54]}} \\
\bottomrule
\end{tabular}
\end{table}

\subsection{Relation Shift and Source-Model Failures}

\subsubsection{Relation-Deviation Score.}
We train SCTransNet \cite{yuan2024sctransnet} on IRSTD-1K and NUDT-SIRST and evaluate it on the unseen
NUAA-SIRST test set. Each relation descriptor is standardized as in
Eq.~\ref{eq:supp_relation_standardization}, using statistics estimated only
from the two source training sets.

Let \(\widehat{\mathbf r}_{i,g}^{\,s}\) denote the standardized subvector of
relation factor \(g\), where
\(\mathcal G=\{\mathrm{sal},\mathrm{mor},\mathrm{clu},\mathrm{sty}\}\).
We define the factor-balanced distance between instances \(i\) and \(q\) as
\begin{equation}
  d(i,q)
  =
  \left[
    \sum_{g\in\mathcal G}
    \frac{
      \left\|
        \widehat{\mathbf r}_{i,g}^{\,s}
        -
        \widehat{\mathbf r}_{q,g}^{\,s}
      \right\|_2^2
    }{|g|}
  \right]^{1/2},
  \label{eq:supp_factor_balanced_distance}
\end{equation}
where division by \(|g|\) gives each relation factor equal weight.

We then build a source reference bank \(\mathcal R_s\) containing equal
numbers of instances from the two source domains. 
The relation-deviation score is the mean distance to the \(k=20\) nearest source instances:
\begin{equation}
  s_i
  =
  \frac{1}{k}
  \sum_{q\in\mathcal N_k(i;\mathcal R_s)}
  d(i,q),
  \; k=20.
  \label{eq:supp_relation_deviation_score}
\end{equation}
A larger \(s_i\) indicates stronger deviation from the source relation
patterns.

\subsubsection{Detection Failure Measure.}
For each target \(T_i\), we measure detection failure by the proportion of its
region not recovered by the binary prediction \(\widehat M\):
\begin{equation}
  \mathrm{FN}_i
  =
  |T_i\setminus\widehat M|,
  \;
  m_i
  =
  \frac{\mathrm{FN}_i}{|T_i|}
  =
  1-
  \frac{|T_i\cap\widehat M|}{|T_i|}.
  \label{eq:supp_target_pixel_miss_rate}
\end{equation}
Here, \(\mathrm{FN}_i\) is the number of missed target pixels and \(m_i\) is
the corresponding miss rate. A larger \(m_i\) indicates a more severe
detection failure.

\begin{figure}[t]
  \centering
  \includegraphics[width=\columnwidth]
  {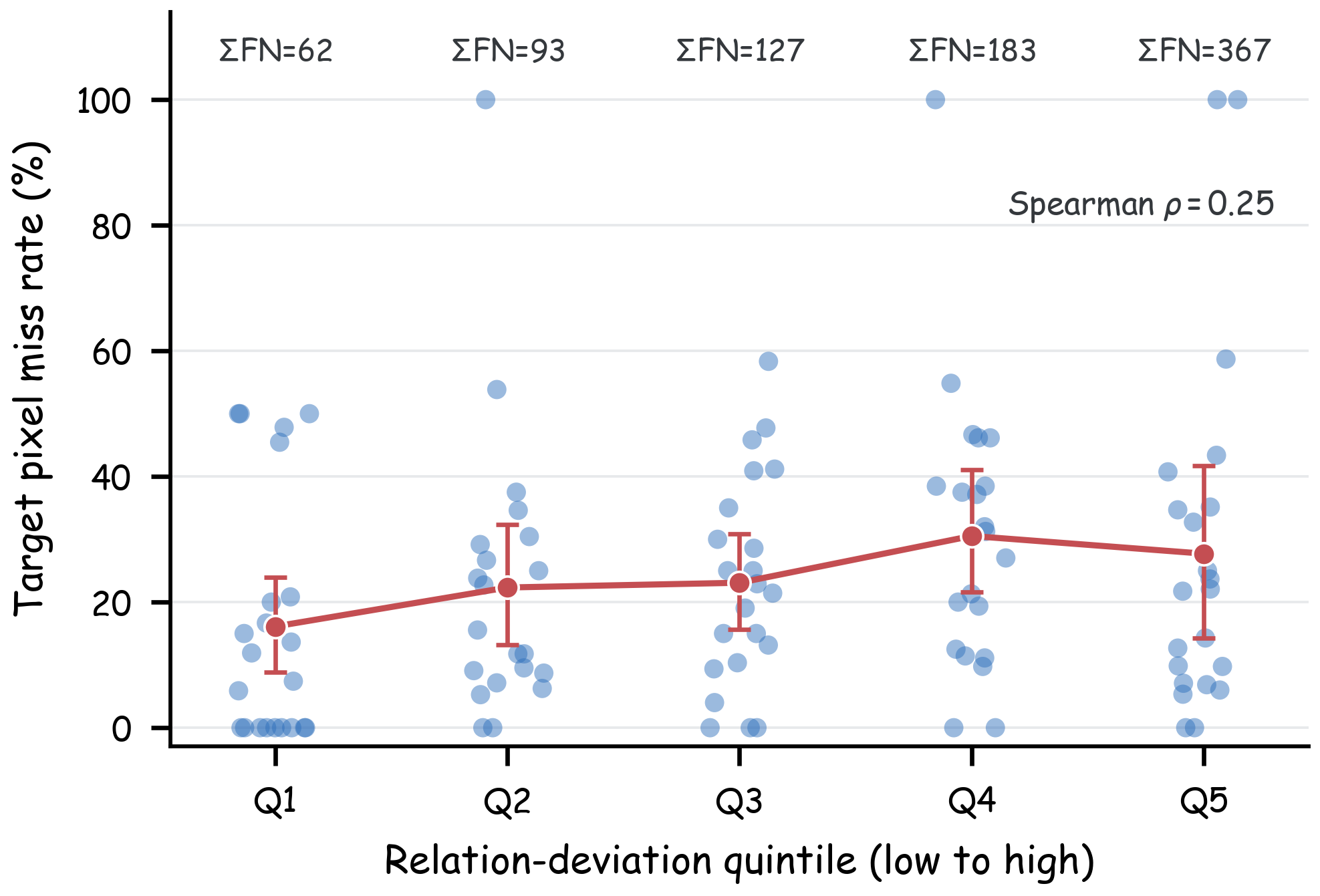}
  \caption{Relation deviation versus target-pixel miss rate on unseen
  NUAA-SIRST. SCTransNet is trained on IRSTD-1K and NUDT-SIRST. Q1--Q5 are
  equal-count groups ordered by deviation. Blue points show individual targets;
  red markers show group means with image-cluster bootstrap 95\% confidence
  intervals; top labels report false-negative pixels.}
  \label{fig:supp_relation_failure_nuaa}
\end{figure}

\subsubsection{Results.}
Across the 108 NUAA-SIRST targets, SCTransNet misses 832 of 3,334 target
pixels, yielding an overall miss rate of \(24.95\%\). We rank these targets by
their relation-deviation scores and divide them into five equal-count groups.
As shown in Fig.~\ref{fig:supp_relation_failure_nuaa}, the mean miss rate
increases from \(16.12\%\) in Q1 to \(30.53\%\) in Q4 and remains high at
\(27.71\%\) in Q5.

At the instance level, the relation-deviation score is positively correlated
with the miss rate (Spearman \(\rho=0.25\), image-cluster bootstrap 95\% CI
\([0.05,0.45]\)). The correlation remains after controlling for normalized
target area (partial Spearman \(\rho=0.26\), 95\% CI \([0.07,0.45]\)) and
changes little for \(k=10,20,50\) (\(\rho=0.24,0.25,0.26\)). These results
indicate that test targets farther from the source-domain target-background
relation patterns generally suffer higher detection miss rates.
% , whileestablishing association rather than causality.

\begin{figure*}[t]
  \centering
  \includegraphics[width=0.98\textwidth]{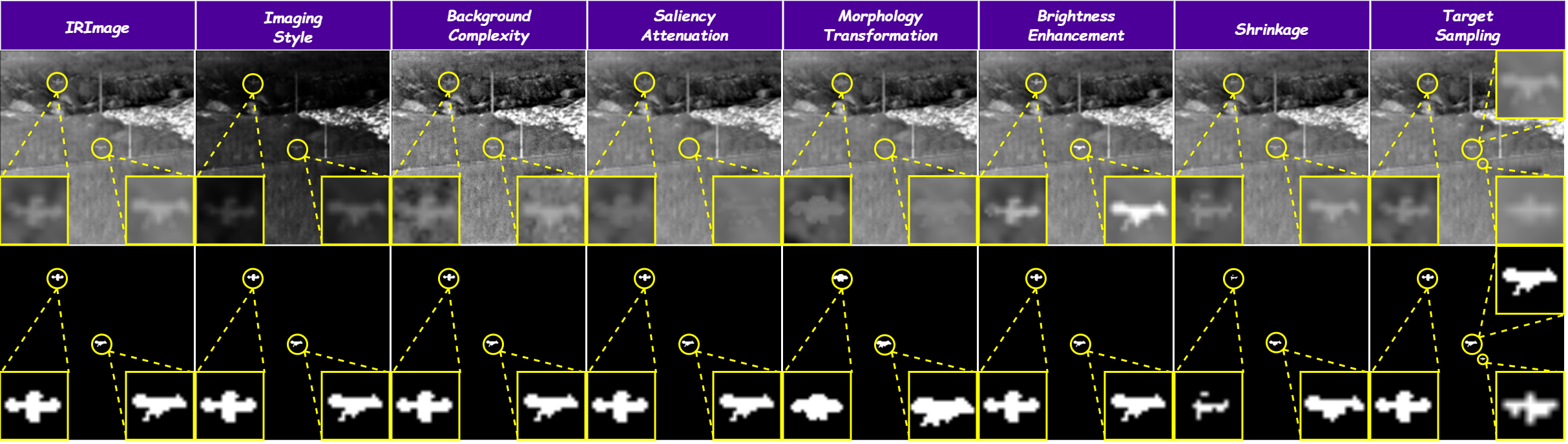}
  \caption{Qualitative examples of the seven TBRI operators. The upper and
  lower rows show the intervened images and corresponding annotations.}
  \label{fig:supp_tbri_examples}
\end{figure*}

\section{Additional Details of Target-Background Relation Intervention}
\label{sec:supp_tbri}

This section specifies the seven operators used in Target-Background Relation
Intervention (TBRI).

\subsection{Operator Overview}
Let \(x\in[0,1]^{H\times W}\) denote a normalized infrared image and
\(y\in\{0,1\}^{H\times W}\) its binary annotation. The target and background
masks are \(M_t=y\) and \(M_b=1-y\), respectively. We organize the seven TBRI
operators into two branches:
\begin{equation}
  \begin{aligned}
  \mathcal{T}
    &=\mathcal{T}_{\mathrm{bg}}\cup\mathcal{T}_{\mathrm{tar}},\\
  \mathcal{T}_{\mathrm{bg}}
    &=\{\mathcal{B}_{\mathrm{sty}},\mathcal{B}_{\mathrm{clu}}\},\\
  \mathcal{T}_{\mathrm{tar}}
    &=\{\mathcal{A}_{\mathrm{sal}},\mathcal{A}_{\mathrm{mor}},
       \mathcal{A}_{\mathrm{bri}},\mathcal{A}_{\mathrm{shr}},
       \mathcal{A}_{\mathrm{sam}}\}.
  \end{aligned}
  \label{eq:supp_tbri_operator_pool}
\end{equation}
The subscripts \(\mathrm{sty}\), \(\mathrm{clu}\), \(\mathrm{sal}\),
\(\mathrm{mor}\), \(\mathrm{bri}\), \(\mathrm{shr}\), and
\(\mathrm{sam}\) denote imaging style, background complexity, saliency
attenuation, morphology transformation, brightness enhancement, shrinkage,
and target sampling, respectively. For each activated sample, one
valid operator is selected from \(\mathcal{T}\). Appearance-only interventions
preserve \(y\), whereas support-changing interventions update it accordingly.
Representative image-annotation pairs are shown in
Fig.~\ref{fig:supp_tbri_examples}.

\subsection{Background Interventions}

The background branch changes the context in which a target is observed while
preserving its annotation.

\subsubsection{Imaging Style.}
This operator jointly perturbs global brightness, contrast, and nonlinear
intensity response:
\begin{equation}
  x_{\mathrm{sty}}
  =
  \operatorname{clip}_{[0,1]}
  \left(
    \operatorname{clip}_{[0,1]}
    [\mu_x+c_s(x-\mu_x)+b_s]^{\gamma_s}
  \right)
  \label{eq:supp_tbri_style}
\end{equation}
where \(\mu_x\) is the mean image intensity, and \(c_s\), \(b_s\), and
\(\gamma_s\) control contrast, brightness, and gamma response, respectively.
The transformation is applied to the entire image, yielding
\((\widetilde{x},\widetilde{y})=(x_{\mathrm{sty}},y)\).

\subsubsection{Background Complexity.}
This operator modifies local fluctuations and structures only within \(M_b\).
Let \(\overline{x}_{15}=\operatorname{AvgPool}_{15}(x)\) and
\(h_5=x-\operatorname{AvgPool}_{5}(x)\). We construct
\begin{equation}
  x_{\mathrm{clu}}
  =
  \overline{x}_{15}
  +
  c_b(x-\overline{x}_{15})
  +
  \lambda_h h_5
  +
  n_w+n_s,
  \label{eq:supp_tbri_background_complexity}
\end{equation}
where \(c_b\) and \(\lambda_h\) control local contrast and high-frequency
enhancement, while \(n_w\) and \(n_s\) denote white and spatially smoothed
noise. The original target is retained:
\begin{equation}
  \widetilde{x}
  =
  M_t\odot x+M_b\odot x_{\mathrm{clu}},
  \;
  \widetilde{y}=y.
  \label{eq:supp_tbri_background_composition}
\end{equation}

\subsection{Target Interventions}

The target branch modifies target intensity, support, or occurrence using the
surrounding background as a local reference. We define the reference ring as
\begin{equation}
  R_b
  =
  \operatorname{Dilate}(M_t;r_{\mathrm{loc}})
  \setminus
  \operatorname{Dilate}(M_t;1),
  \label{eq:supp_tbri_local_ring}
\end{equation}
after excluding all target pixels. Its mean and standard deviation are denoted
by \(\mu_{R_b}\) and \(\sigma_{R_b}\). If the ring is too small, these
statistics are estimated from \(M_b\).

\subsubsection{Saliency Attenuation.}
This operator reduces the target--background contrast while preserving the
target support. We measure the mean and peak contrasts as
\begin{equation}
  \begin{aligned}
  \mathrm{CNR}
    &= \frac{|\mu_{M_t}-\mu_{R_b}|}
            {\sigma_{R_b}+\epsilon},\\
  \mathrm{PSNR}_{t}
    &= \frac{|p_t-\mu_{R_b}|}
            {\sigma_{R_b}+\epsilon},
  \end{aligned}
  \label{eq:supp_tbri_target_contrast}
\end{equation}
where \(p_t\) is the maximum target intensity for a bright target and the
minimum target intensity for a dark target.

Given the sampled contrast levels \(\tau_c\) and \(\tau_p\), the retained
contrast factor is
\begin{equation}
  \begin{aligned}
  \alpha_0
    &=
    \min\left\{
      \frac{\tau_c}{\mathrm{CNR}+\epsilon},
      \frac{\tau_p}{\mathrm{PSNR}_{t}+\epsilon}
    \right\},\\
  \alpha
    &=
    \operatorname{clip}_{[\alpha_{\min},\alpha_{\max}]}
    (\alpha_0).
  \end{aligned}
  \label{eq:supp_tbri_saliency_alpha}
\end{equation}
The target residual relative to the local background is then scaled by
\(\alpha\):
\begin{equation}
  \begin{aligned}
  \widetilde{x}
    &= M_b\odot x
       +M_t\odot
       \left[\mu_{R_b}+\alpha(x-\mu_{R_b})\right],\\
  \widetilde{y}
    &= y.
  \end{aligned}
  \label{eq:supp_tbri_saliency_response}
\end{equation}
A smaller \(\alpha\) moves the target intensity closer to its local background,
thereby producing stronger saliency attenuation.

\subsubsection{Morphology Transformation.}
This operator changes target shape using a sampled offset set
\(\mathcal{O}_m\), including elongated, curved, broken, asymmetric, and
block-like patterns:
\begin{equation}
  M_t'
  =
  \mathbb{I}
  \left[
    \max_{(\Delta u,\Delta v)\in\mathcal{O}_m}
    \operatorname{Shift}
    (M_t;\Delta u,\Delta v)
    >0
  \right].
  \label{eq:supp_tbri_morphology_mask}
\end{equation}
Target intensities are propagated with the same offsets and averaged in
overlapping regions to obtain \(x_{\mathrm{mor}}\). Image and annotation are
then updated jointly:
\begin{equation}
  \widetilde{x}
  =
  (1-M_t')\odot x+M_t'\odot x_{\mathrm{mor}},
  \;
  \widetilde{y}=M_t'.
  \label{eq:supp_tbri_morphology_composition}
\end{equation}

\subsubsection{Brightness Enhancement.}
This operator strengthens the positive target response relative to the local
background:
\begin{equation}
  x_{\mathrm{bri}}
  =
  \mu_{R_b}
  +
  g_b|x-\mu_{R_b}|,
  \label{eq:supp_tbri_brightness}
\end{equation}
where \(g_b>1\) is a sampled gain. The target support is unchanged:
\(\widetilde{x}=M_b\odot x+M_t\odot x_{\mathrm{bri}}\) and
\(\widetilde{y}=y\).

\subsubsection{Shrinkage.}
This operator retains target pixels with strong local contrast and high
centrality. For each \(p\in M_t\), we first compute its absolute residual from
the local-background mean and normalize it within the target:
\begin{equation}
  \delta(p)=|x(p)-\mu_{R_b}|,
  \;
  \widehat{\delta}(p)
  =
  \frac{\delta(p)-\delta_{\min}}
       {\delta_{\max}-\delta_{\min}+\epsilon},
  \label{eq:supp_tbri_shrink_residual}
\end{equation}
where \(\delta_{\min}\) and \(\delta_{\max}\) are the minimum and maximum of
\(\delta(p)\) over \(M_t\), respectively. Each target pixel is then ranked by
\begin{equation}
  s(p)
  =
  0.65\,\widehat{\delta}(p)
  +
  0.35
  \left(
    1-\frac{\|p-c_t\|_2}{d_{\max}+\epsilon}
  \right),
  \label{eq:supp_tbri_shrink_score}
\end{equation}
where \(c_t\) is the target centroid and \(d_{\max}\) is the maximum distance
from \(c_t\) to a target pixel. Given a sampled keep ratio \(\eta\), the number
of retained pixels is
\begin{equation}
  n_{\mathrm{keep}}
  =
  \min\!\left\{
    |M_t|,
    \max\!\left[1,\operatorname{round}(\eta|M_t|)\right]
  \right\}.
  \label{eq:supp_tbri_shrink_keep_number}
\end{equation}
The \(n_{\mathrm{keep}}\) pixels with the highest scores define the reduced
support \(M_t'\). Removed pixels are filled by
\begin{equation}
  x_{\mathrm{fill}}
  =
  0.65\,\operatorname{AvgPool}_{k}(x)
  +
  0.35\,\mu_{R_b}.
  \label{eq:supp_tbri_shrink_fill}
\end{equation}
With \(M_r=M_t-M_t'\), the resulting pair is
\begin{equation}
  \widetilde{x}
  =
  (1-M_r)\odot x+M_r\odot x_{\mathrm{fill}},
  \;
  \widetilde{y}=M_t'.
  \label{eq:supp_tbri_shrink_composition}
\end{equation}

\subsubsection{Target Sampling.}
This operator samples an annotated source-domain target, rescales it, and
places it at a valid background location \(\ell\). Let \(x_s^\ell\),
\(M_s^\ell\), and \(\overline{M}_s^\ell\) denote its response, binary support,
and soft blending mask, respectively. The resulting pair is
\begin{equation}
  \widetilde{x}
  =
  (1-\overline{M}_s^\ell)\odot x
  +
  \overline{M}_s^\ell\odot x_s^\ell,
  \;
  \widetilde{y}
  =
  y\lor M_s^\ell.
  \label{eq:supp_tbri_random_target_sampling}
\end{equation}
Candidate locations overlapping existing targets are rejected, and the
sampled support is added to the annotation.

\section{Implementation Details and Hyperparameter Analysis}
\label{sec:supp_implementation}

This section supplements the implementation details omitted from the main
paper, covering the model architecture and the operations used for
Poincar\'e-ball relation modeling. We then conduct hyperparameter analyses.

\subsection{Implementation Details}
\subsubsection{Encoder.}
The encoder contains five feature levels with spatial resolutions
\(\{256,128,64,32,16\}\), each containing \(32\) channels. 
Let \(X_1=x\).
At level \(i\), a ResBlock\cite{he2016deep} produces
\begin{equation}
  F_i=\mathcal{R}_i(X_i),
  \;
  X_{i+1}=\operatorname{MaxPool}_{2}(F_i),
  \label{eq:supp_encoder_levels}
\end{equation}
where pooling is omitted after the final level. 
Each block first projects its input as
\(\widetilde X=\delta(\operatorname{BN}[\operatorname{Conv}_{3}(X)])\)
and then applies a \(5\times5\)--\(3\times3\) residual branch:
\begin{equation}
  \begin{aligned}
  \widehat X
    &=
    \delta\!\left(
      \operatorname{BN}[\operatorname{Conv}_{5}(\widetilde X)]
    \right),\\
  \mathcal{R}_i(X)
    &=
    \delta\!\left(
      \widetilde X+
      \operatorname{BN}[\operatorname{Conv}_{3}(\widehat X)]
    \right).
  \end{aligned}
  \label{eq:supp_encoder_resblock}
\end{equation}
Here, \(\widehat X\) is the intermediate branch feature and \(\delta\) denotes
ReLU.

\subsubsection{Decoder.}
The decoder reconstructs target representations from coarse to fine. At level
\(i\), the preceding target feature \(\widehat T_{i+1}\) is bilinearly
upsampled and fused with the adapted encoder feature \(A_i\):
\begin{equation}
  B_i=
  \operatorname{Conv}_{1}
  \left([A_i,\operatorname{Up}(\widehat T_{i+1})]\right).
  \label{eq:supp_decoder_fusion}
\end{equation}
At the coarsest level \(I=5\), we set \(B_I=A_I\).

\begin{figure}[t]
  \centering
  \includegraphics[width=0.90\columnwidth]
  {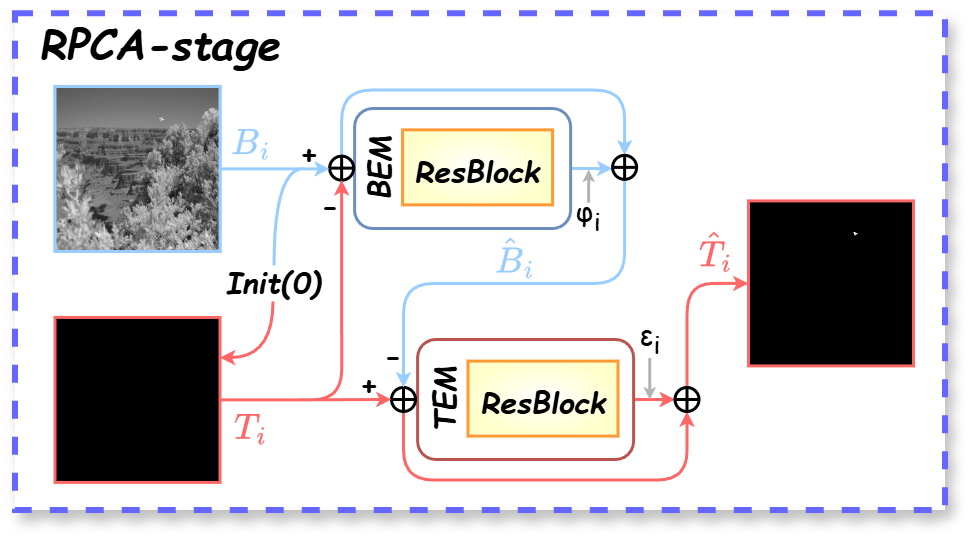}
  \caption{Architecture of the RPCA stage within the RPCABlock.}
  \label{fig:RPCA-Stage}
\end{figure}

Each decoder level contains one RPCABlock, as shown in
Fig.~\ref{fig:RPCA-Stage}. Following \cite{liu2026rpcassm, wu2024rpcanet}, the block
performs one background update followed by one target update.

With \(T_i=0\), the two updates are
\begin{equation}
  \begin{aligned}
  \widehat B_i
    &=\mathcal{R}_{B,i}(B_i-T_i)
      +\varphi_i(B_i-T_i),\\
  \widehat T_i
    &=\mathcal{R}_{T,i}(T_i-\widehat B_i)
      +\varepsilon_i(T_i-\widehat B_i).
  \end{aligned}
  \label{eq:supp_rpca_update}
\end{equation}
Here, \(\mathcal{R}_{B,i}\) and \(\mathcal{R}_{T,i}\) denote two 
ResBlock. The learnable coefficients \(\varphi_i\) and
\(\varepsilon_i\) are initialized to \(0.01\). 
% During training, all five
% outputs \(\{\widehat T_i\}_{i=1}^{5}\) are resized to \(256\times256\) and
% jointly supervised.

\subsubsection{Poincar\'e Mapping and Distance.}
We detail the two Poincar\'e-ball operations used by HRM. Following
Ganea et al.~\cite{ganea2018hyperbolic}, each relation token \(u_i\) is mapped
from the tangent space at the origin to \(z_i\) in the Poincar\'e ball:
\begin{equation}
  z_i=\operatorname{Exp}_{0}^{c}(u_i)
  =
  \tanh\!\left(\sqrt{c}\lVert u_i\rVert_2\right)
  \frac{u_i}{\sqrt{c}\lVert u_i\rVert_2}.
  \label{eq:supp_poincare_expmap}
\end{equation}
The target and background anchors follow the notation of the main paper:
\begin{equation}
  a_t=\operatorname{Exp}_{0}^{c}(\rho),
  \;
  a_b=\operatorname{Exp}_{0}^{c}(-\rho),
  \label{eq:supp_poincare_anchors}
\end{equation}
where \(a_t\) and \(a_b\) are the target and background anchors, respectively,
and \(\rho\) controls their scale. For \(a\in\{a_t,a_b\}\), the geodesic
distance is
\begin{equation}
  d_c(z_i,a)
  =
  \frac{2}{\sqrt c}
  \operatorname{artanh}
  \left(
    \sqrt c\,
    \lVert(-z_i)\oplus_c a\rVert_2
  \right).
  \label{eq:supp_poincare_distance}
\end{equation}
Here, \(\oplus_c\) denotes M\"obius addition. The relation score used in the
main paper is therefore
\begin{equation}
  s_i=d_c(z_i,a_b)-d_c(z_i,a_t).
  \label{eq:supp_relation_score}
\end{equation}
We use \(c=1\) and \(\rho=0.05\) in the default configuration.

% \subsubsection{Hard-Background Mining and Loss Weights.}
% In addition to the target and background terms defined in the main paper, the
% implementation penalizes the background tokens with the largest relation
% scores. Let \(\Omega_h\) contain the top
% \(\lceil\eta|\Omega_b|\rceil\) background tokens ranked by \(s_i\). We use
% \(\eta=0.1\) and define
% \begin{equation}
%   \mathcal{L}_{h}
%   =
%   \frac{1}{|\Omega_h|}
%   \sum_{i\in\Omega_h}
%   [m+s_i]_+,
%   \qquad
%   \mathcal{L}_{\mathrm{hyp}}
%   =
%   \mathcal{L}_{t}
%   +
%   \mathcal{L}_{b}
%   +
%   0.5\mathcal{L}_{h}.
%   \label{eq:supp_hard_background_loss}
% \end{equation}
% The complete training objective is
% \begin{equation}
%   \mathcal{L}
%   =
%   \mathcal{L}_{\mathrm{IoU}}
%   +\lambda_{\mathrm{hyp}}\mathcal{L}_{\mathrm{hyp}}
%   +\lambda_{\mathrm{lb}}\mathcal{L}_{\mathrm{lb}}
%   +\lambda_{\mathrm{nc}}\mathcal{L}_{\mathrm{nc}},
%   \label{eq:supp_training_objective}
% \end{equation}
% where
% \(\lambda_{\mathrm{hyp}}=\lambda_{\mathrm{lb}}
% =\lambda_{\mathrm{nc}}=0.1\).
% HMA uses four experts at each of the five feature levels, and its learnable
% residual scales are initialized to \(0.01\). The remaining optimization and
% preprocessing settings follow the main paper.

\subsection{Hyperparameter Ablation}

\subsubsection{Number of Experts.}
We examine the number of experts \(E\) in HMA under the same
NUDT-SIRST + IRSTD-1K \(\rightarrow\) NUAA-SIRST setting used for the ablation
studies in the main paper. We vary \(E\in\{1,2,4,8\}\) while keeping all other
training and architectural settings unchanged. The default configuration uses
\(E=4\).

\begin{table}[t]
\centering
\caption{Ablation of the number of experts \(E\) on the
NUDT-SIRST + IRSTD-1K \(\rightarrow\) NUAA-SIRST setting. All other
components and training settings are kept unchanged. Results are reported in
\(mIoU\) (\%), \(F\)-measure (\%), \(P_d\) (\%), and \(F_a\)
(\(10^{-6}\)).}
\label{tab:supp_num_experts}
\scriptsize
\setlength{\tabcolsep}{2.8pt}
\resizebox{\columnwidth}{!}{%
\begin{tabular}{@{}c@{\hspace{7pt}}cccccc@{}}
\toprule
& \multicolumn{4}{c}{Detection performance}
& \multicolumn{2}{c}{Complexity} \\
\cmidrule(lr){2-5}\cmidrule(l){6-7}
\(E\)
& \(mIoU\uparrow\)
& \(F\uparrow\)
& \(P_d\uparrow\)
& \(F_a\downarrow\)
& Params (M)
& FLOPs (G) \\
\midrule
1 & 75.74 & 86.19 & 99.07 & 15.25 & 1.06 & 7.99 \\
2 & 77.57 & 87.37 & 99.23 & 12.20 & 1.08 & 8.13 \\
\(4\)\rlap{\hspace{0.05em}\textsuperscript{\(\star\)}}
& 78.58 & 88.00 & 99.56 & 13.10 & 1.13 & 8.42 \\
8 & 76.54 & 86.71 & 99.12 & 8.07 & 1.22 & 9.01 \\
\bottomrule
\end{tabular}%
}
\end{table}

Performance improves as \(E\) increases from 1 to 4, with \(E=4\) achieving
the highest \(mIoU\), \(F\)-measure, and \(P_d\). Increasing \(E\) to 8
further reduces \(F_a\), but degrades the overlap-based metrics and increases
the computational cost. We therefore use \(E=4\) as the default setting,
which provides the best overall accuracy--complexity trade-off.

\subsubsection{Intervention Probability and Anchor Scale.}
We further examine the activation probability \(p\) of TBRI and the anchor
scale \(\rho\) of HRM. We vary one hyperparameter at a time while retaining
the default value of the other and keeping all remaining settings unchanged.
For TBRI, \(p\) controls the proportion of training samples exposed to the
intervention operators. For HRM, \(\rho\) controls the separation between the
target and background anchors in the tangent space. The default configuration
uses \(p=0.5\) and \(\rho=0.05\).

\begin{table}[H]
\centering
\caption{Hyperparameter sensitivity under the NUDT-SIRST + IRSTD-1K
\(\rightarrow\) NUAA-SIRST setting. Each parameter is varied independently;
a star indicates the default. \(mIoU\), \(F\), and \(P_d\) are reported
in \%, and \(F_a\) in \(10^{-6}\).}
\label{tab:supp_p_rho_sensitivity}
\footnotesize
\setlength{\tabcolsep}{4pt}
\renewcommand{\arraystretch}{1.02}
\begin{tabular*}{\columnwidth}{@{\extracolsep{\fill}}lcccc@{}}
\toprule
Value & \(mIoU\uparrow\) & \(F\uparrow\)
& \(P_d\uparrow\) & \(F_a\downarrow\) \\
\midrule
\multicolumn{5}{l}{\textit{TBRI: activation probability \(p\)}} \\
\cmidrule(lr){1-5}
\(0.25\) & 77.47 & 87.30 & 99.37 & 15.61 \\
\(0.50\)\rlap{\hspace{0.05em}\textsuperscript{\(\star\)}}
& 78.58 & 88.00 & 99.56 & 13.10 \\
\(0.75\) & 76.77 & 86.86 & 98.67 & 21.18 \\
\(1.00\) & 77.60 & 87.39 & 99.07 & 13.48 \\
\addlinespace[2pt]
\multicolumn{5}{l}{\textit{HRM: anchor scale \(\rho\)}} \\
\cmidrule(lr){1-5}
\(0.01\) & 74.79 & 85.57 & 96.29 & 31.59 \\
\(0.03\) & 76.77 & 86.86 & 98.07 & 24.05 \\
\(0.05\)\rlap{\hspace{0.05em}\textsuperscript{\(\star\)}}
& 78.58 & 88.00 & 99.56 & 13.10 \\
\(0.07\) & 75.52 & 86.05 & 97.96 & 26.56 \\
\(0.10\) & 76.62 & 86.76 & 98.14 & 31.41 \\
\bottomrule
\end{tabular*}
\end{table}

Table~\ref{tab:supp_p_rho_sensitivity} shows that TBRI performs best at the
moderate activation probability \(p=0.50\), achieving the highest \(mIoU\),
\(F\)-measure, and \(P_d\), together with the lowest \(F_a\). Reducing \(p\)
to \(0.25\) weakens the overall performance, whereas increasing it to \(0.75\)
or \(1.00\) provides no further improvement. In particular, \(p=0.75\)
increases \(F_a\) to \(21.18\).

HRM exhibits a similar intermediate optimum. The default scale
\(\rho=0.05\) achieves the best result across all four metrics. Both smaller
scales (\(0.01\) and \(0.03\)) and larger scales (\(0.07\) and \(0.10\))
reduce the detection accuracy and increase false alarms. These results support
the default configuration \(p=0.50\) and \(\rho=0.05\).

\end{document}